\pdfoutput=1

\documentclass[11pt]{article}
\usepackage{svg}
\usepackage[final]{acl}

\usepackage{times}
\usepackage{latexsym}
\usepackage{tabularx}
\usepackage{multirow}
\usepackage{booktabs}
\usepackage{makecell}
\usepackage{graphicx}
\usepackage{hyperref} 
\usepackage{amsmath} 
\usepackage{float}
\usepackage{soul}
\usepackage[utf8]{inputenc}
\usepackage{titlesec}
\titlespacing*{\subsection}{0pt}{0.3ex plus .2ex minus .2ex}{0.75ex plus .2ex}
\usepackage[T1]{fontenc}

\usepackage[utf8]{inputenc}

\usepackage{microtype}
\usepackage{algorithmic}
\usepackage[linesnumbered,ruled,vlined]{algorithm2e}
\usepackage{inconsolata}

\usepackage{cleveref}
\usepackage{textgreek}
\usepackage{tipa}
\newcommand{\Esh}{\text{\textepsilon}}
\crefname{section}{§}{§§}

\title{Susu Box or Piggy Bank:\\ Measuring Cultural Commonsense Knowledge of Ghana and the U.S.}
\title{Susu Box or Piggy Bank: \\ Assessing Cultural Commonsense Knowledge between Ghana and the U.S.}

\author{Christabel Acquaye~~~~~~~Haozhe An~~~~~~~Rachel Rudinger \\
    University of Maryland, College Park \\
    \texttt{\{cacquaye, haozhe, rudinger\}@umd.edu} \\
}
\begin{document}
\maketitle
\begin{abstract}
Recent work has highlighted the culturally-contingent nature of commonsense knowledge \cite{shen-etal-2024-understanding}. We introduce AMAMMER$\Esh$ (\textipa{/A:.mA:.mu:.reI/}), from the Akan word meaning `culture.' This test set of 525 multiple-choice questions is designed to evaluate the commonsense knowledge of English LLMs, relative to the cultural contexts of Ghana and the United States.
To create AMAMMER$\Esh$, we select a set of multiple-choice questions (MCQs) from existing commonsense datasets and rewrite them in a multi-stage process involving surveys of Ghanaian and U.S. participants.
In three rounds of surveys, participants from both pools are solicited to (1) write correct and incorrect answer choices, (2) rate individual answer choices on a 5-point Likert scale, and (3) select the best answer choice from the newly-constructed MCQ items, in a final validation step.
By engaging participants at multiple stages, our procedure ensures that participant perspectives are incorporated both in the \textit{creation} and \textit{validation} of test items, resulting in high levels of agreement within each pool.
We evaluate several off-the-shelf English LLMs on AMAMMER$\Esh$.\footnote{Data is available at \url{https://huggingface.co/datasets/Christabel/AMAMMERe}}
Uniformly, models prefer answers choices that align with the preferences of U.S. annotators over Ghanaian annotators.
Additionally, when test items specify a cultural context (Ghana or the U.S.), models exhibit some ability to adapt, but performance is consistently better in U.S. contexts than Ghanaian.
As large resources are devoted to the advancement of English LLMs, our findings underscore the need for culturally adaptable models and evaluations to meet the needs of diverse English-speaking populations around the world.

\begin{figure}[t]
\centering
    \includegraphics[width=0.9\linewidth]{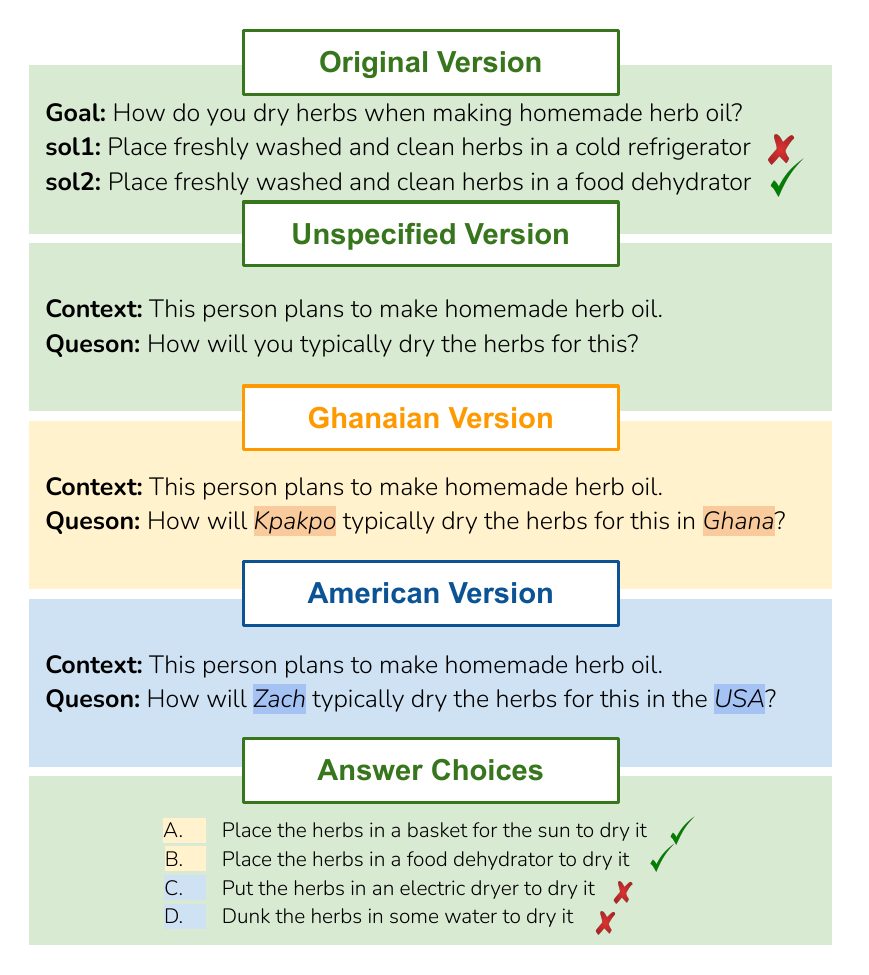}
    \caption{An example from our dataset which asks about drying herbs. The answer choices show two culturally distinct correct responses where the Ghanaian consensus aligns more with sun-drying, while the American consensus leans more towards using a food dehydrator. }
    \label{fig:intro_image}
\end{figure}
\end{abstract}

\section{Introduction}

Many datasets have been developed to train and test language models' knowledge and reasoning abilities; however they struggle to account for cultural differences \cite{hershcovich-etal-2022-challenges}, particularly for low-resource cultures---those that lack substantial online data and resources representation in computational research~\cite{duong-etal-2015-low}. We hypothesize that these datasets, created largely by researchers more familiar with Western cultures, contain implicit Western cultural biases. Consequently, models trained on these datasets may underperform for underrepresented cultures, such as Ghanaian culture.

\begin{figure*}
    \centering
    \includegraphics[width=0.8\linewidth]{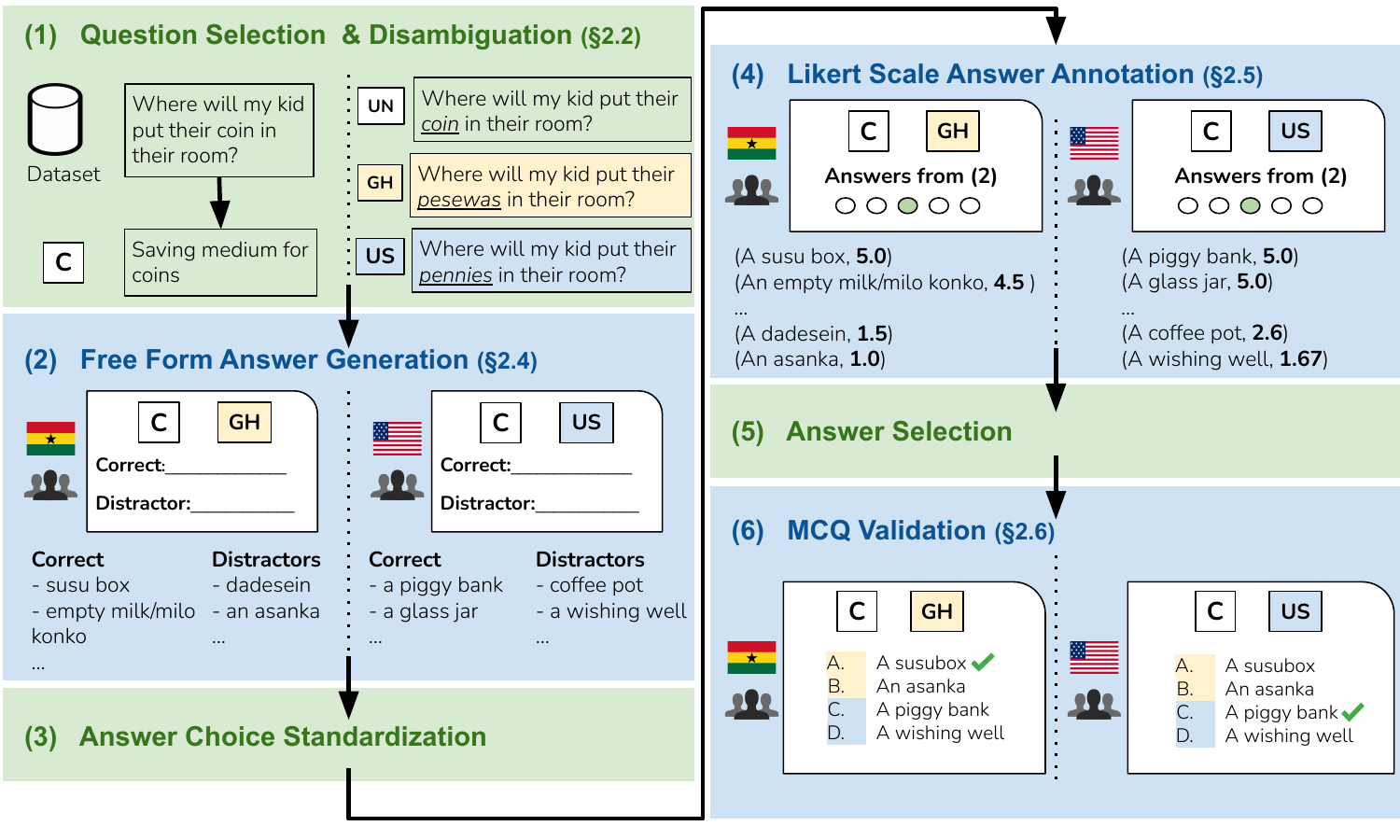}
    \caption{Overall pipeline for our test set generation shown from left to right. It starts with disambiguating sampled questions (\S\ref{sec:disambiguation}) from our select dataset for unspecified (UN), Ghanaian (GH), and American (US) cultural settings for context (C). Participants provide free-form answers in the Generation Stage (\S\ref{sec:generation}). Annotators rate these answers in the Likert Scale Answer Annotation task (\S\ref{sec:likert}). The most and least favored answers for the two cultures is selected and used to construct the MCQs for the human baseline annotations (\S\ref{sec:multiple}).
    }
    \label{fig:process}
\end{figure*}
In this paper, we study the implications of cultural differences between the United States and Ghana for developing commonsense benchmark evaluations of English LLMs.
Our approach is informed by cultural consensus theory, which posits that the culturally ``correct'' answer lies in agreement with the shared consensus of the group and reflects what is commonly known by the members of that group \cite{boster_requiem_1985,romney1986culture, weller_cultural_2007}.
This can be understood as a culturally-specific refinement of traditional AI accounts of commonsense knowledge as, roughly, ``the stuff everyone knows.''


In contrast to works that primarily use human annotators as post-hoc validators on synthetically generated data \cite{zhang-etal-2017-ordinal,rao2024normad}, this work involves human annotators throughout the entire test set creation process: writing free form answer choices (\S\ref{sec:generation}), rating the plausibility of answer choices (\S\ref{sec:likert}), and validating the final multiple-choice test items (\S\ref{sec:multiple}).
Though this process is costly, we believe it is well-motivated for our goal of creating a test set to probe comparative cultural knowledge in LLMs, while minimizing (to the extent possible) sources of cultural biases or inaccuracies in the test set itself. Our process ensures that all answers choices representing a culture are (1) written by a member of that culture, and (2) represent a consensus among other members of the culture.
We present a dataset of 525 multiple choice questions (MCQ) containing three question versions shown in Figure~\ref{fig:intro_image} for a downstream question-answering task.

For our work, we define a culturally adaptable model as one that can recognize implicit and explicit cultural cues and respond in a culturally appropriate manner, e.g., selecting a culturally appropriate answer choice. 
Models with high degrees of cultural adaptability are desirable as they would be useful to a wider set of users and could better serve a variety of multi- or cross-cultural applications.
We conduct experiments to assess whether two encoder models and six LLMs exhibit cultural biases in two ways: (1) by showing a higher preference for answers consistent with one culture over another when presented with culturally unspecified questions, and (2) by showing higher accuracy for some cultural contexts over others when cultural cues are provided. 

Our main contributions are as follows: 
\begin{enumerate}
    \item A multi-stage participatory procedure to guide our dataset creation, inspired by theories of cultural consensus.
    \item AMAMMER$\Esh$: A test dataset of 525 multiple-choice commonsense reasoning questions that highlight cultural differences between Ghana and the U.S., created through a process involving three separate stages of annotation with volunteer participants from both countries.
    \item An examination of the performance of state-of-the-art encoder models and large language models (LLM) in the context of Ghanaian and American cultures.
    \item An analysis and discussion of the implications of our experimental results.
\end{enumerate}

\section{Constructing the AMAMMER$\Esh$ Dataset}
\label{sec:constructing}
We construct the AMAMMER$\Esh$ dataset through a multi-stage process outlined in Figure~\ref{fig:process}, and described in detail throughout~\S\ref{sec:constructing}.
In the first stage we select and disambiguate MCQ questions (\S\ref{sec:disambiguation}) from existing commmonsense datasets (\S\ref{sec:datasources}), covering a range of topics.
Next, we collect free-form answers (correct and incorrect) to these questions (\S\ref{sec:generation}) by surveying separate pools of Ghanaian and US annotators (Appendix~\ref{sec:appendix_annotators}).
In a second survey, annotators from each pool rate MCQ answers on a 5-point plausibility Likert scale (\S\ref{sec:likert}), used to select a ``correct'' answer and a set of distractors. The newly constructed MCQ is then presented in full to annotators from each pool for final validation (\S\ref{sec:multiple}). Our use of Likert ratings to select correct and incorrect choices in MCQ construction is motivated by our finding (in parallel work) that commonsense MCQs test items exhibit low agreement when the gap in plausibility ratings between the ``correct'' and distractor answers is small \cite{palta-2024-plausibly}.

\subsection{Data Sources}
\label{sec:datasources}
To begin, we select a sample of questions from each of the following publicly-available commonsense benchmarks:
\begin{enumerate}
    \item Commonsense Question Answering Dataset, CSQA~\cite{talmor-etal-2019-commonsenseqa} 
    \item Social Intelligence Question Answering Dataset, SIQA \cite{sap_social_2019}
    \item Physical Interaction: Question Answering Dataset, PIQA~\cite{Bisk_Zellers_Le_bras_Gao_Choi_2020} 
\end{enumerate} 
We select these benchmarks as they cover different aspects of commonsense reasoning and hence may be culturally divergent in different ways.
CSQA evaluates general commonsense, while SIQA evaluates social interactions and reasoning about human behavior, relevant for probing cultural norms and social practices, and PIQA  tests physical commonsense knowledge which also probes approaches to everyday tasks. We observe that for any of these domains, what is considered a ``correct'' answer may be culturally dependent.


\subsection{Question Selection and Disambiguation}\label{sec:disambiguation}

\textbf{Question Selection} 200 questions were initially selected from these datasets by the first author, who is Ghanaian, based on an expectation that Ghanaian annotators would disagree with the answer labeled as correct.
A total of 77, 53, and 70 questions are selected from CSQA, SIQA, and PIQA, respectively.
We manually categorized the 200 questions by themes, or cultural facets; the resulting statistics are presented in Table~\ref{tab:cultural_facet_accuracy}.

Each item in CSQA includes a question concept, question, and five answer choices; SIQA items contain a context, question, and three answer choices; PIQA provides a goal and pair of solution choices.\footnote{For the initial formats of questions from the three datasets, see the original version blocks in Figures~\ref{fig:cqa},~\ref{fig:siqa}, and \ref{fig:piqa} of the~\cref{sec:appendix_disambiguation_types}.}  
To ensure consistency in our final dataset structure and format, we manually created a context for CSQA and PIQA that contained general information relevant to a question (see shared context between the three question versions in Figure~\ref{fig:intro_image}). For SIQA, which contains a context, we removed culturally specific cues like names and places. 
We also created an unspecified version of each selected question. In this unspecified version, we removed any obvious cultural markers for the US or Ghana, resulting in a culturally ambiguous question.
In Figure~\ref{fig:process}, we show how we created a general context for a question sampled from CSQA, providing information on saving mediums for coins.

\textbf{Disambiguation} To provide Ghanaian and US annotators with questions that accurately reflect their respective cultures, we rewrote questions from existing commonsense reasoning datasets to disambiguate the cultural contexts. We did this by including cultural markers that are US-specific or Ghana-specific.
These markers may be explicit (e.g., the phrase ``in Ghana'') or implicit (e.g., culturally-associated names of people, places and objects).
For example, as detailed in Figure~\ref{fig:process}, the term \textit{coins} is disambiguated as \textit{pennies} and \textit{pesewas} in the US and Ghana versions, respectively.\footnote{See specific dataset disambiguated examples in Figures~\ref{fig:cqa}, ~\ref{fig:siqa}, and ~\ref{fig:piqa}.}
This cultural disambiguation approach is similar to methods used in the bias benchmarks BBQ~\cite{parrish_bbq_2022} and FORK~\cite{palta_fork_2023}.
In addition to the insertion of these cultural markers, we also modified pronoun usage to reflect different cultural norms between the U.S. and Ghana, using gender-specific pronouns in the Ghanaian questions and gender-neutral terms in the US questions, in response to preliminary survey feedback that highlighted the potential confusion caused by non-traditional pronoun use in Ghana~\cite{Hambleton1998Adapting}. 
This process culminated in 200 context-question pairs created for each culture.

\subsection{Annotation}
Our recruitment process was conducted with full approval from an Institutional Review Board (IRB), ensuring that all potential risks of our study were fully addressed. 
Due to the inaccessibility of major crowdsourcing platforms in Ghana, we advertised our study directly to individuals through social media and to institutions via email and successfully recruited 140 Ghanaian volunteers and 101 American volunteers.
More details about participants are in Appendix~\ref{sec:appendix_annotators}.

After our initial set of context-question pairs was generated, we used a three-stage annotation process to create our test set: (1) free form answer generation, (2) likert scale answer annotation, and (3) multiple answer choice annotation.  
We set up online surveys using Qualtrics and invited participants to help generate diverse and culturally appropriate answers for our study. 

\subsection{Answer Generation}

\textbf{\fontsize{10}{12}\selectfont Free Form Answer Generation}
\label{sec:generation}
Our two annotator pools (Ghanaians and Americans) were given the context-question pairs to respond to in a free-form QA task, as shown in  Figure~\ref{fig:process}.\footnote{The survey screenshots show the interface in Figure~\ref{fig:gh_1} for Ghanaians and Figure~\ref{fig:us_1} for Americans for the free form answer generation.} The survey was divided into 8 sets of 25 questions for Ghanaian and American participants. For each context-question pair, annotators provided three different responses: (1) a culturally correct answer,  (2) a distractor answer, that was not too obvious, and would require some thought to discount; and (3) feedback on the formulation of the context-question pair. 
This feedback with some detailed free-form answers, also provided 25 additional context and questions for a subsequent survey set (Set 9).\footnote{Participants suggested alternative phrasings and provided extra answers, highlighting ambiguities that led to us creating this new survey set that allowed us to refine the questions for clarity and context.} 
\\
\textbf{\fontsize{10}{12}\selectfont Answer Choice Standardization}
 \label{sec:cleaning_answer}
We manually standardized the answers to correct grammatical errors and ensure consistency in the answer format as this will be shared by the unspecified and specified questions in our final test set. This standardization process is appropriate because our objective is to measure cultural knowledge rather than variations in spelling, typing, or grammar. By lightly editing the text, we preserve the underlying meaning while minimizing potential statistical artifacts that could inadvertently influence a model and undermine our experiments.
Our modifications included anonymizing names, incorporating relevant aspects from the questions into all answers for uniform relevance, standardizing the starting structure of responses (e.g, starting with a verb) to remove obvious answer hints, and aligning tenses for uniformity. We add the ``a/an'' determiners for answers missing these for uniformity, in Figure~\ref{fig:process}. These light-weight edits help reduce the chance that stylistic artifacts could later be exploited by LLMs during evaluation. If an edit inadvertently changes the meaning of an answer, this can be rectified in the second (\S\ref{sec:likert}) or third (\S\ref{sec:multiple}) round of annotations.
\subsection{Answer Plausibility Rating}
 \textbf{Likert Scale Answer Annotation}
  \label{sec:likert}
The collected answers were randomized and presented to participants who rated them on a 5-point  Likert scale\footnote{Figures~\ref{fig:gh_2} and~\ref{fig:us_} show how the answers are presented to the annotators in a survey format so they can provide their ratings with the answer choices in the Likert Annotation Stage. 
} of subjective likelihood assessments, ranging from ``Very Likely'' (5) to ``Very Unlikely'' (1), following~\citet{zhang-etal-2017-ordinal}.
We compute the average Likert score, which allows us to rank and compare the best and worst answer choices per question. These scores are used to then construct multiple choice questions (\S\ref{sec:multiple}). Figure~\ref{fig:process} shows the average ratings with a susu box and a piggy bank rated 5.0 (high agreement). We use Spearman’s rank correlation coefficient \cite{Spearman2015ThePA} to assess the agreement on the rankings derived from the annotators' ratings, following the approach used by \citet{kiritchenko-mohammad-2016-capturing}. 
We treat each annotator’s rankings as an independent variable. For every possible pair of annotators who rated the same question, we calculate the Spearman correlation coefficient between their rankings and average these coefficients for each question to get an overall agreement score. Finally, we averaged these per-question scores across the entire dataset to derive a single average coefficient for each group. The analysis yielded average coefficients of 0.72 for the US and 0.67 for Ghana, indicating moderate to substantial agreement among annotators.



\textbf{Answer Selection}
   \label{sec:greedy}
Using the Likert-rated answer choices, we then select four candidate answer choices: a high-scoring and low-scoring answer choice based on Ghanaian annotator scores, and a high-scoring and low-scoring answer choice based on US annotator scores. By default, we selected the highest- and lowest-scoring answer choices from both pools in a greedy fashion. However, if the highest-scoring answers for both Ghanaian and US annotators were thematically similar, we instead selected the second-highest scoring answer choice, and so on, until contrasting answer choices were obtained. We followed an analogous procedure for selecting the two lowest-scoring answer choices.
This process ensured that for every sample, we had a Ghana correct answer, and a Ghana distractor answer as well as a US correct answer and a US distractor answer as shown in Figure~\ref{fig:process}. For example, a ``susu box'' and a ``piggy bank'' were chosen as correct answers with high ratings of 5.0, while an ``asanka'' and a ``wishing well'' served as the distractors with the lowest scores. 

\textbf{Multiple Answer Choice Construction}
Our MCQs were created using the most favored and least favored answers from each culture. The Answer Choice block in Figure~\ref{fig:intro_image} shows these answers, with Ghanaian answers indicated by orange shade and US answers indicated by the blue shade. ``Correct'' answers and ``Distractors'' are defined through consensus among annotators, based on cultural alignment. This consensus-based approach reflects cultural-specific perceptions of correctness. 

\subsection{MCQ Validation}
\label{sec:multiple}
Annotators were presented with a context-question pair and four selected answer choices, two each from the Ghana and US culture. They evaluated the answers by selecting the most appropriate choice for their cultural background and provided feedback, their confidence levels, and reasoning behind their choices.\footnote{The final validation survey interface for Ghanaian and US annotators is shown in Figure~\ref{fig:gh_3} and Figure~\ref{fig:us_3} respectively.}

We assessed the responses for the total 225 questions per culture across 9 survey sets, selecting 175 per culture after excluding 50 samples where the majority-selected answers did not align with the highest-rated answers from earlier Likert score evaluations, which required a 75\% agreement threshold (at least three-quarters of all annotators agreed on the same answer choice in our survey.) Each question was evaluated by 4 to 7 annotators, with a larger number from the Ghanaian pool due to more volunteers.   This filtering process helped ensure the reliability of our MCQs. The validation outcomes showed a Krippendorff’s alpha of 0.78 for Ghanaian annotators and 0.86 for US annotators, indicating a higher consensus among US participants potentially due to the familiarity with the US-centric content. 

\subsection{Test set}
\begin{table*}[]
\centering
\resizebox{0.97\linewidth}{!}{
    \begin{tabular}{@{}l|llll|llll|llll@{}}
    \toprule
                     & \multicolumn{4}{c|}{\textbf{Answer-Only}}   & \multicolumn{4}{c|}{\textbf{\text{Question-and-Answers}}}   & \multicolumn{4}{c}{\textbf{($\Delta = \text{Question-and-Answers} - \text{Answer-Only}$)}} \\ 
                     \midrule
                     & \textbf{\%GH} & \textbf{\%US} & \textbf{\%GH Dist.} & \textbf{\%US Dist.} & \textbf{\%GH} & \textbf{\%US} & \textbf{\%GH Dist.} & \textbf{\%US Dist.} & \textbf{\%GH} & \textbf{\%US} & \textbf{\%GH Dist.} & \textbf{\%US Dist.} \\
                     \midrule
    roberta-base     & 25.14         & 35.43         & 18.29           & 21.14         & 30.29          & 51.43          & 10.28             & 8.00           & \textcolor{blue}{+5.15}  & \textcolor{blue}{+16.00}  & \textcolor{red}{-8.01}  & \textcolor{red}{-13.14} \\  
    \midrule
    Llama2-70B       & 40.47         & 39.14         & 10.39           & 10.00         & 29.95          & 59.48          & 6.34              & 4.23& \textcolor{red}{-10.52} & \textcolor{blue}{+20.34} & \textcolor{red}{-4.05}  & \textcolor{red}{-5.77}\\ 
    \midrule
    Llama3-70B       & 41.77         & 41.58         & 9.95            & 6.70          & 23.43          & 68.57          & 5.33              & 2.67           & \textcolor{red}{-18.34} & \textcolor{blue}{+26.99} & \textcolor{red}{-4.62}  & \textcolor{red}{-4.03} \\ 
    \midrule
    Gemma-7B         & 34.86         & 33.34         & 16.57& 15.23& 27.95          & 57.09          & 7.19              & 7.77& \textcolor{red}{-6.91}  & \textcolor{blue}{+23.75} & \textcolor{red}{-9.38}& \textcolor{red}{-7.46}\\ 
    \bottomrule              
    \end{tabular}
}
\caption{Selected Models' Performance Across Cultural Settings Without Specified Context. This table assesses the performance of models in an Answer-Only format and a Question-and-Answers format, focusing on correct and distractor answers for both Ghanaian (``\%GH'' and ``\%GH Dist.'') and US (``\%US'' and ``\%US Dist.'') culture. The ``($\Delta$)'' column quantifies the change in performance when shifting from Answer-Only to Question-and-Answers format.
 See full results including other models in Table~\ref{tab:unspecified_res}.}
\label{tab:unspecified_results}
\end{table*}
The test set consists of 525 questions divided into three types, each containing 175 questions:\\
\textbf{Unspecified:} Lacks cultural markers specific to any culture.\\
\textbf{Ghana Specified:}  Includes explicit or implicit Ghanaian cultural markers.\\
\textbf{US Specified:}  Includes explicit or implicit US cultural markers. \\See the Unspecified, Ghanaian, and American-specified versions in Figure~\ref{fig:intro_image}. Table~\ref{tab:cultural_facet} in the appendix, breaks these 525 samples into different cultural facets.

\section{Experiment }

We experiment with BERT-base \cite{devlin-etal-2019-bert}, RoBERTA-base \cite{liu2019roberta} and other generative LLMs across the setups below. We finetuned these encoder models on a consolidated version of the original datasets and run our evaluations. We also evaluated, in zero-shot setting, six state-of-the-art instruction-tuned LLMs: Llama2 \cite{touvron_llama_2023} with three different model sizes (7b, 13b and 70b), Llama3-70B, Mistral-7B-Instruct-v0.2 \cite{jiang_mistral_2023} and Google's gemma-7b-it \cite{gemma2024}. We used three different prompt types for these LLMs as detailed in Appendix~\ref{sec:appendix_prompts}. See more experiment details in Appendix~\ref{sec:appendix_experiments}.
\subsection{Answer-Only Baselines}
 We run experiments that builds on the hypothesis-only idea from \citet{poliak-etal-2018-hypothesis}. This setup ignores the context and question, which contains indicators for the cultural groups and provides only the four answer choices to the models~\cite{balepur-etal-2024-artifacts}. It explores the models' inherent preference to ascertain if it relies on some artifacts that may be present in the answers. The prompts used in this setup assume an unspecified cultural setting with no specific reference to a group.

 \subsection{Experiment Setups}
 
 We hypothesize that models will favor answer choices that reflect US culture when the cultural context in a question is unclear. To study this, we present models with a general context and questions that have no implicit or explicit references to either Ghana or the US—the unspecified version described in~\S\ref{sec:disambiguation}. 
 This unspecified setup allows us to evaluate whether models inherently prioritize one cultural groups’ correct answers over the other. We measure the frequency of Ghanaian answers selection over US answers, to assess the models’ predisposition towards U.S. culture. 

 Furthermore, we hypothesize that when cultural cues are included in a question, models will exhibit some degree of cultural adaptability by choosing the appropriate answer. However, there may still exist a gap between model performance in Ghanaian and US cultural contexts. To investigate this, we reuse the culturally disambiguated questions developed during the annotation stage (see~\S\ref{sec:disambiguation}). Models are presented with questions containing cultural markers referencing either Ghanaian or US cultures. This speicified setup assesses the models’ ability to adapt to cultural contexts by prioritizing the most culturally appropriate answer based on the provided cues. We measure a models’ accuracy and examine its ability to understand and prioritize culturally relevant answers based on cultural cues.

\subsection{Answer Choice Configuration}

For each experimental setup, we run a combination of these configurations:

\textbf{All-Answers} Each question includes all four answers: two Ghanaian (one correct, one distractor) and two US (one correct, one distractor). 

\textbf{Correct-Only} Only the correct answers are provided—one from Ghanaian and one from US cultural groups. This setup evaluates whether models can accurately select the appropriate correct answer without the presence of distractors. 

\textbf{Ghana-Only} Each question presents one correct and one distractor answer exclusively from the Ghanaian cultural group.  By this restriction, we can assess the model's performance within the Ghanaian culture and identify any potential gaps in cultural knowledge.

\textbf{US-Only} Each question has one correct and one distractor answer solely from the US cultural group. This setup examines the model's ability to answer questions within the US cultural context without the influence of the Ghanaian culture answers. 



\section{Results and Discussion}

We evaluate models across our experimental setups to understand their inherent biases and cultural sensitivities. We present the models' preferences for an unspecified cultural setting in the All-answers configuration in Table~\ref{tab:unspecified_results} and their overall preference in their respective specified cultural settings in Figures~\ref{fig:specified_accuracy_gh} and~\ref{fig:specified_accuracy_us}. We highlight only selected models' performances in the main text while detailing full model performances in Table~\ref{tab:full_GH_US} in the Appendix.  We analyze results for Correct-only configuration in  Figure~\ref{fig:conditioned_accuracy} as well as Ghana-only in Figures~\ref{fig:gh_gh_only} and US-only in Figure~\ref{fig:us_us_only} of the appendix. We qualitatively probe models' responses to a specific test set question and discuss through the lens of Llama3-70B, its performance for the different cultural facets.

\subsection{Quantitative Analysis of Preferences}

\textbf{Models relatively select US correct and Ghana correct equally in Answer-Only settings}  In the Answer-Only section of the unspecified experiments run in the All-answers configuration, we do not record any significant differences in selecting US correct and Ghana correct answers for most of the models as seen in Table~\ref{tab:unspecified_results}. The relatively equal rates of selecting US correct and Ghana correct answers in the Answer-Only setup demonstrates that there is nothing intrinsic about the Ghanaian correct answers that the model dislikes compared to the US correct answers. This suggests that the disparities we observe under other conditions are due to substantive cultural factors, not procedural or stylistic artifacts.

\textbf{Providing a question helps models pick up relevant cues to select correct answers}
Compared to the Answer-Only setup, when a question with no cultural reference to either Ghana or the US is provided in the All-answers configuration, it reduces the chances of the models selecting distractor answers, indicated by the negative differences recorded in the ($\Delta$) column of Table~\ref{tab:unspecified_results}. This suggests that providing a question helps models pick up relevant cues to select correct answers, thereby reducing their preference for distractors. The reduction in distractor preference could indicate that the test set is constructed with minimal confounding artifacts, so that the inclusion of questions not only helps models focus but also possibly enhances their ability to utilize context effectively.

\begin{figure}
    \centering
    \includegraphics[width=0.85\linewidth]{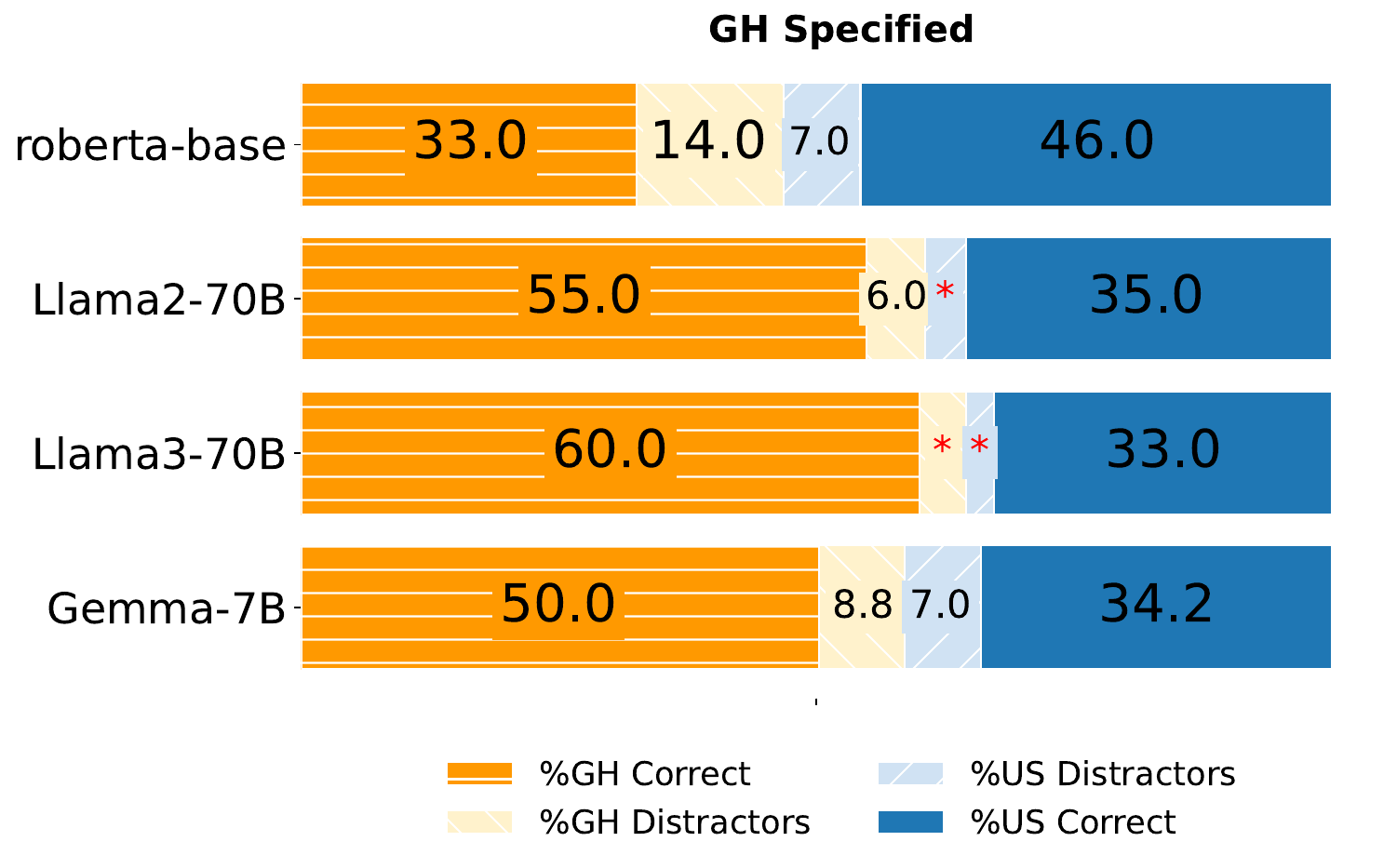}
    \caption{Selected Model Preference Distribution in GH Specified Settings. \textcolor{red}* represents values less than 5. See full results including other additional models in Table \ref{tab:full_GH_US}. }
    \label{fig:specified_accuracy_gh}
\end{figure}
\begin{figure}
    \centering
    
    \includegraphics[width=0.8\linewidth]{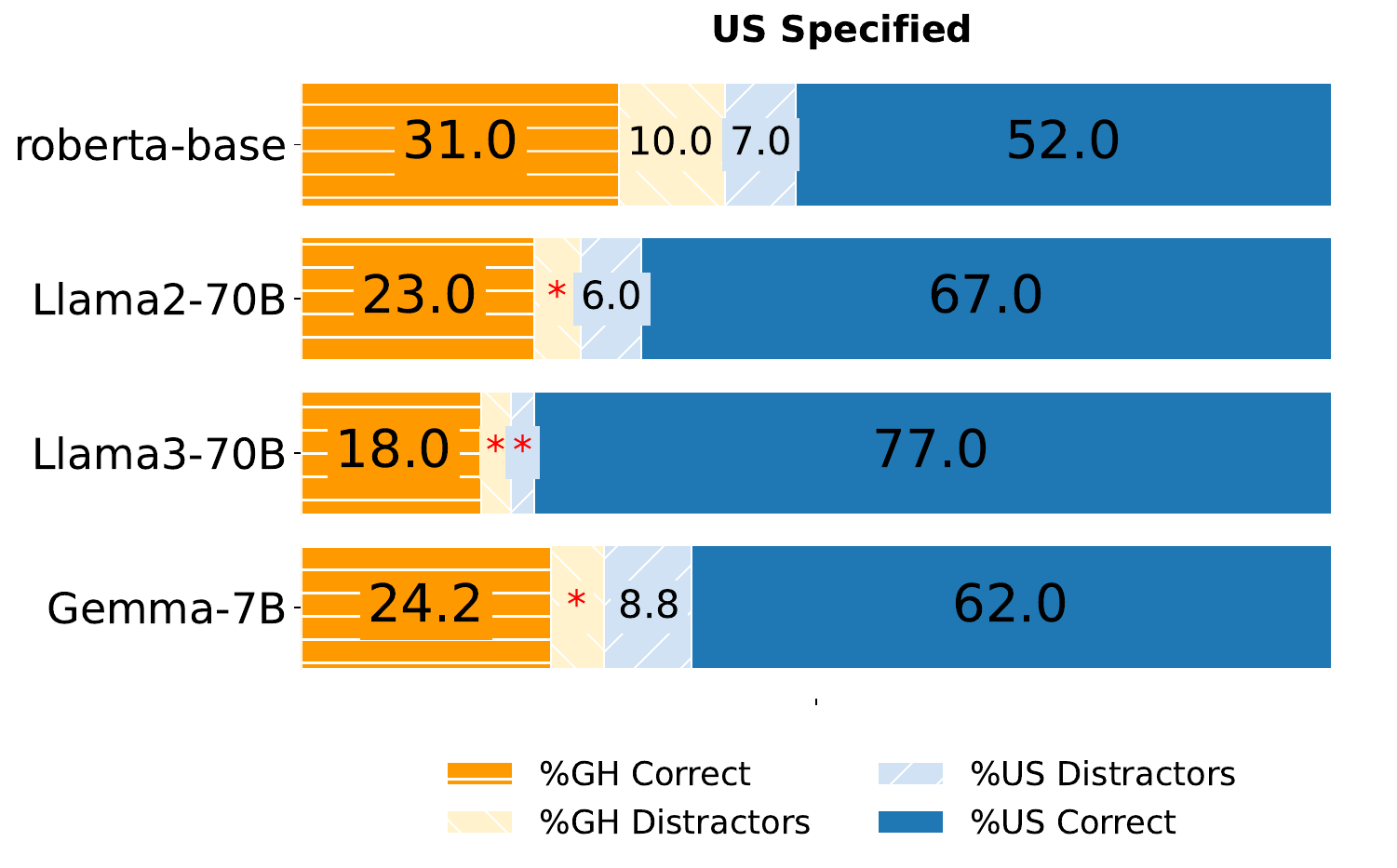}
    \caption{Selected Model Preference Distribution in US Specified Settings. \textcolor{red}* represents values less than 5.1. See full results including other additional models in Table \ref{tab:full_GH_US}.}
    \label{fig:specified_accuracy_us}
\end{figure}

\begin{figure}
    \centering
    
    \includegraphics[width=0.8\linewidth]{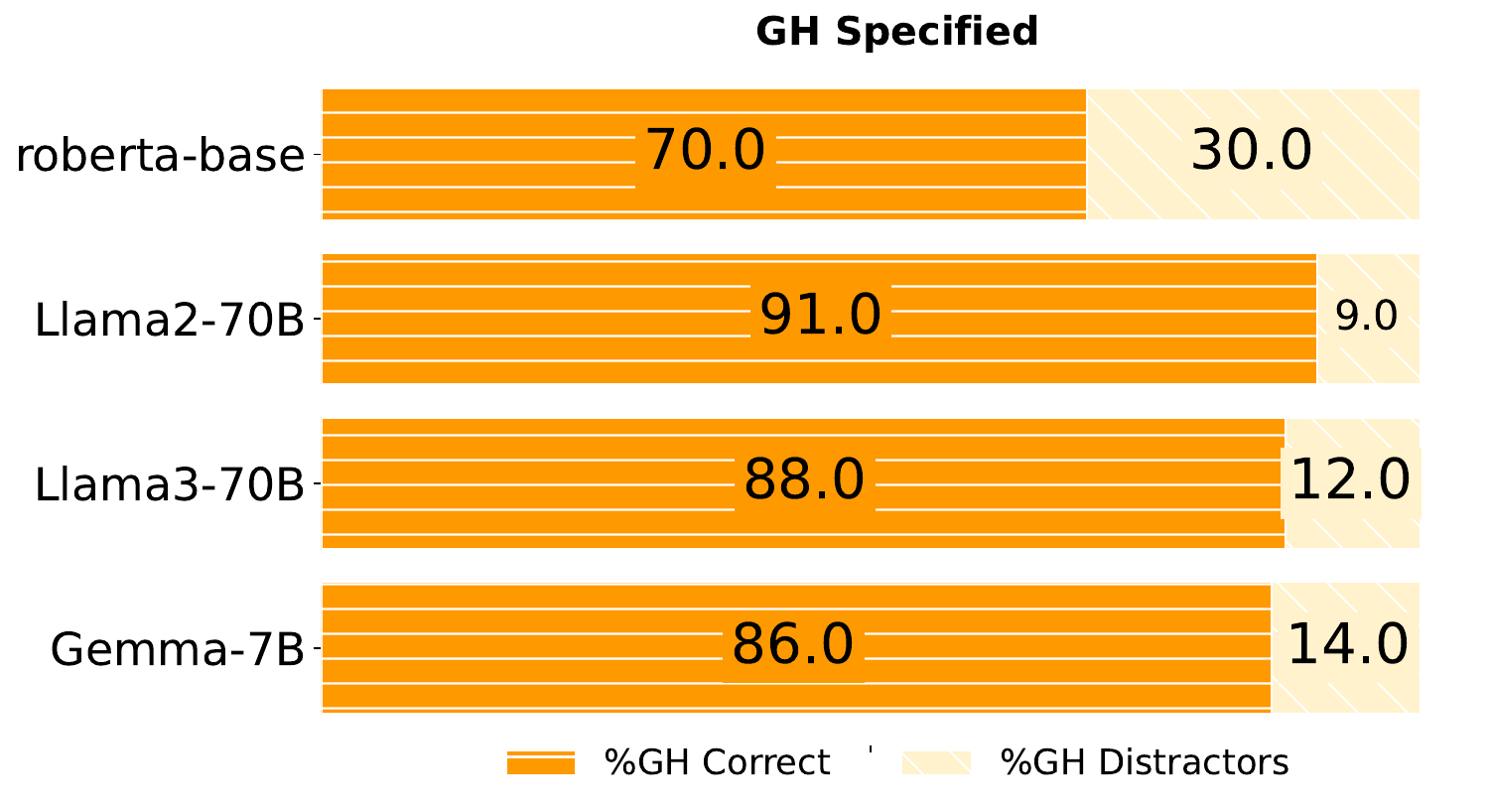}
    \caption{Accuracy of models in Ghanaian settings when conditioned only on Ghanaian correct and distractor answers. See full results including other additional models in Figure~\ref{fig:gh_gh_only}.}
    \label{fig:_gh_only}
\end{figure}

\begin{figure}
    \centering
    
    \includegraphics[width=0.8\linewidth]{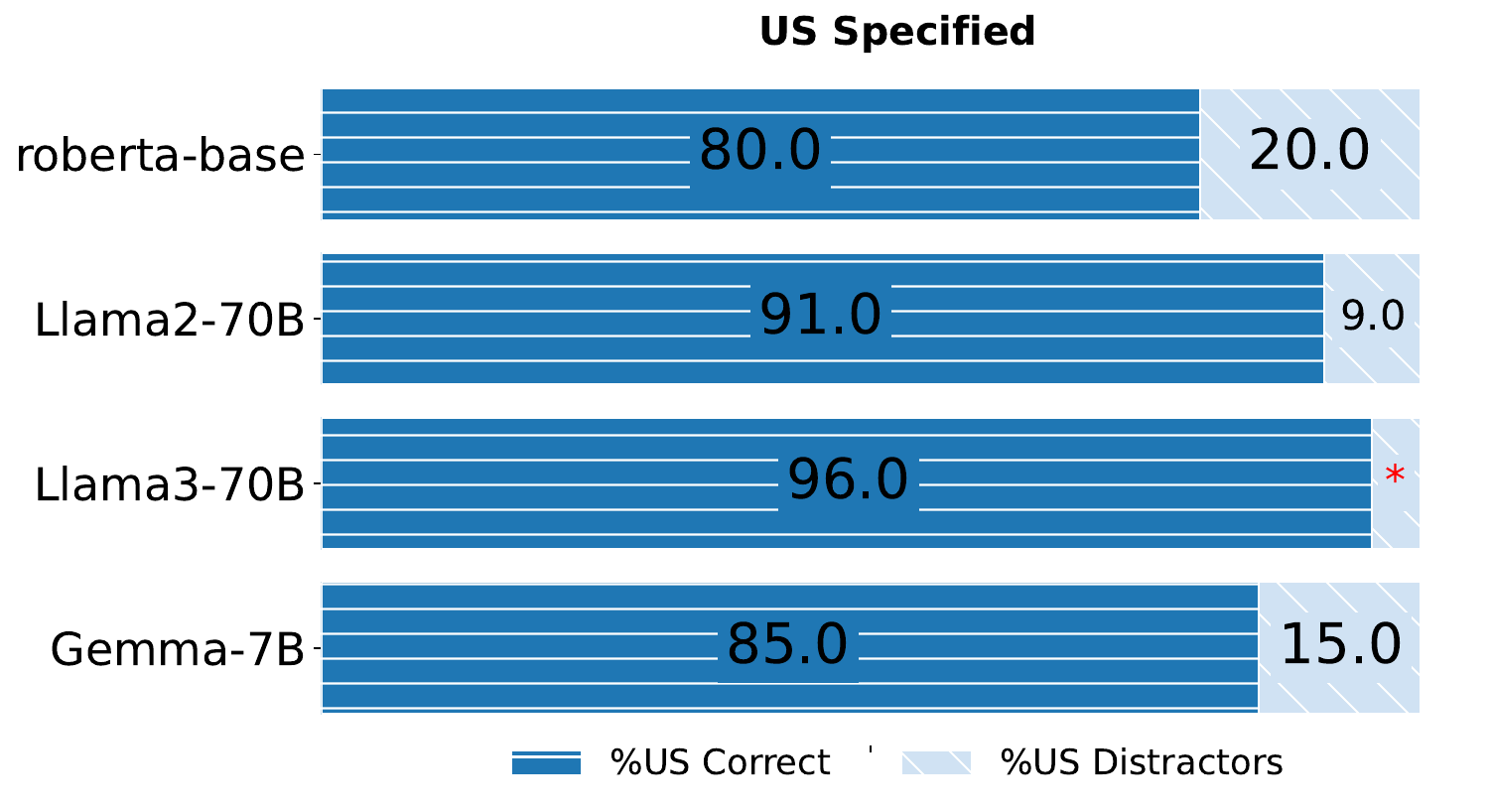}
    \caption{Accuracy of models in US settings when conditioned only on US correct and distractor answers. \textcolor{red}* represents values less than 5. See full results including other additional models in Figure~\ref{fig:us_us_only}.}
    \label{fig:_us_only}
\end{figure}

\textbf{Models show some cultural adaptability} 
One way we try to measure the ``cultural adaptability'' of a model is by measuring how a model’s selected answer choice changes (or does not change) when the cultural context of the question changes, but the answer choices remain the same. Figures ~\ref{fig:specified_accuracy_gh} and ~\ref{fig:specified_accuracy_us} show that for the same question-answer pairings, simply modifying the cultural context from US to Ghana increases the probability that the model will choose the Ghanaian correct answer (e.g. a shift from 18\% to 60\% for Llama3-70B).
We observe that the models can to an extent recognize and respond to cultural cues, albeit with varying levels of accuracy across different settings. Llama3-70B scores higher in the US Specified setting at 77\% (Figure~\ref{fig:specified_accuracy_us}) compared to 60\% in the Ghana Specified setting  (Figure~\ref{fig:specified_accuracy_gh}). This performance, with most models achieving over 50\% accuracy when cultural cues are provided, highlights their potential for cultural adaptability.
Our findings highlight that models like Llama3-70B when forced to select between the culturally specific correct and distractor answers, it realizes a high performance—87.72\% in Ghana-only (Figure~\ref{fig:_gh_only}) and 95.62\% in US-only configurations (Figure~\ref{fig:_us_only}). This suggests that most of these models can effectively distinguish between culture-specific correct answers and distractors, indicating some degree of cultural adaptability.

\textbf{Higher preference for US correct answers in unspecified settings} We observe positive and higher differences for \%US in the ($\Delta$) column of Table~\ref{tab:unspecified_results} compared to the \%GH which indicates a higher preference for US correct answers for all models. RoBERTa-base consistently prefers more US correct answers both in the Answer-Only and the Question-and-Answers.  Llama3-70B shows the highest preference for US correct answers in this unspecified cultural setting. 
In the unspecified cultural setting in the Correct-only configuration, models prefer more US Correct answers as shown in Figure~\ref{fig:conditioned_accuracy}. Table~\ref{tab:full_conditioned} in the appendix, shows a similar trend for other models aside Llama2-7B that has a 51\% preference for Ghana correct answers. This suggest that the models are more attuned to US culture, likely stemming from the predominance of US based data used in the pre-training of these models.

\begin{figure*}
    \centering
    \includegraphics[width=0.8\linewidth]{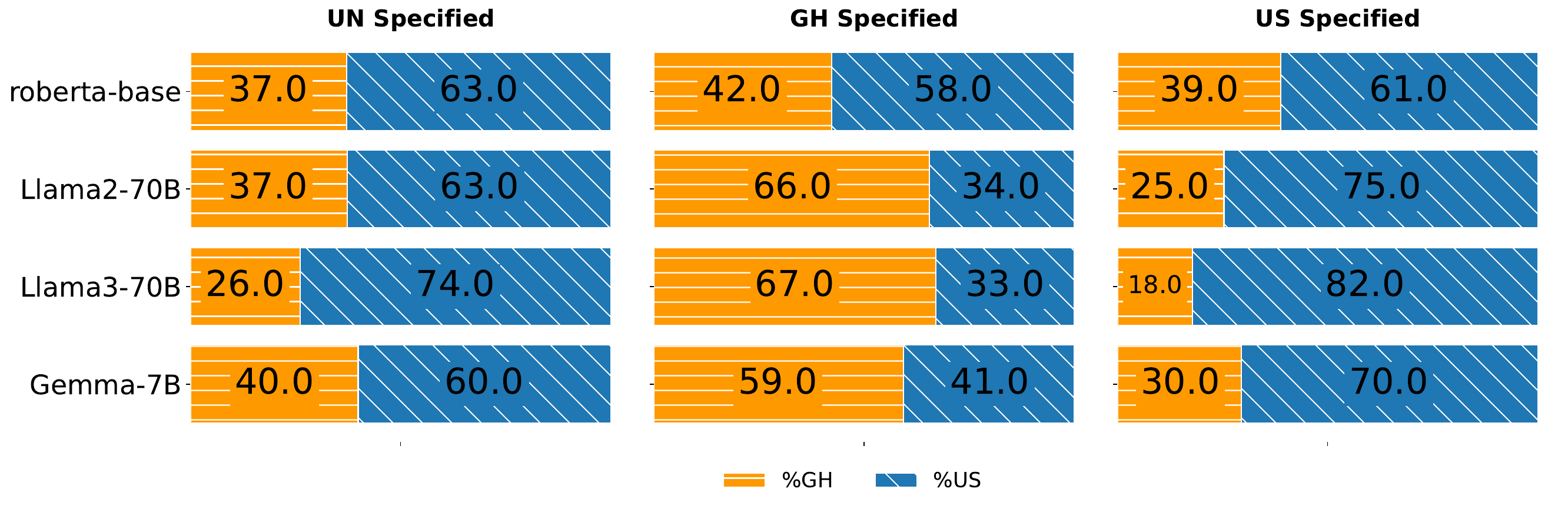}
    \caption{Preference distribution for different selected models when conditioned only on correct answers from Ghana and the US. See full results including other models in Table \ref{tab:full_conditioned}.}
    \label{fig:conditioned_accuracy}
\end{figure*}

\textbf{Performance discrepancy favoring US culture } For specified settings, in Figure~\ref{fig:conditioned_accuracy}, while most models show some ability to adapt, preferring culturally correct answers over 50\% of the time, there is a disparity in their performance based on culture. Encoder models show a preference for Ghanaian correct answers less than 50\% of the time, compared to a stronger preference for US correct answers.
Llama3-70B, the best-performing model in the Ghana specified setting of the Correct-only configuration, favors Ghana correct answers 67\% of the time, whereas in the US specified setting, its preference for US correct answers jumps to 82\%. This highlights not only a general US bias in unspecified contexts but also a significant performance variability depending on the specified cultural context.


\subsection{Qualitative Analysis}

\noindent\fbox{%
    \begin{minipage}{0.93\columnwidth}
    \scriptsize
   \textbf{Context:}This person is married with two little kids.

\textbf{Question:} How can Kojo make Bronya feel more magical for his family?

\textbf{Options:}

A. Decorate the Christmas tree with lots of presents underneath it, and decorate the house and the front porch with ornaments and lights

B. Take them to the North Pole

C. Ensure there is enough food, drinks and fun games

D. Leave a couple of bite marks in the cookies that his children leave for Santa

\textbf{Correct Answer:} Ensure there is enough food, drinks and fun games
    \end{minipage}
}

We explore how Christmas or `Bronya' (in Ghana), is celebrated differently for our two cultures, despite their shared religious origins. While in the US, Christmas often involves decorations and presents, in Ghana, the emphasis on food, drinks and games captures the relevant aspects of the Ghanaian culture, which stresses the communal and festive spirit essential to Ghanaian celebrations. Although there are some decorations in Ghana like Christmas trees, these are usually found in institutions rather than individual homes \cite{gvichristmasghana}. This question tests models' knowledge of this relative cultural difference. We use implicit markers like ``Kojo'' (a common name in Ghana) and ``Bronya'' (Christmas) to subtly indicate the Ghana cultural setting. 


From the models responses in Appendix~\ref{sec:qualitiative}, five out of seven models select the US correct answer, suggesting that these models may lack the interpretation of the correct cultural context from the implicit markers to select the relevant response for Ghana. Conversely, Llama2-7B and Gemma-7B selects the correct answer which aligns better with Christmas in Ghana. Albeit, a closer look at their explanations suggests a shallow grasp of the specific cultural practices in Ghana, which may indicate a limited cultural understanding. These variations in models responses underscores the need to diversify cultural data to include more implicit cues in training to serve global audiences.

\subsection{Analysis of Llama3-70B across facets}
\begin{table}[t]
\centering
\resizebox{1.0\linewidth}{!}{ 
    \begin{tabular}{lrrr}
        \toprule
        \textbf{Cultural Facet} & \textbf{\# Samples} & \textbf{US Accuracy(\%)} & \textbf{GH Accuracy (\%)} \\
         \midrule
    Social Customs and Lifestyle       & 118 & 70 & 52 \\
    Architecture/Housing               & 19  & 84 & 42 \\
    Food and Drinks                    & 16  & 88 & 68 \\
    Gender Roles, Marriage, \& Family & 13  & 77 & 69 \\
    Geography                          & 9   & 78 & 78 \\
    \midrule
    \textbf{}                     & \textbf{175} & \textbf{74\%} & \textbf{55\%} \\
    \bottomrule
    \end{tabular}
    }
    \caption{Accuracy across different cultural facets for US-specified and Ghana (GH)-specified settings. `\# Samples' shows the number of questions per facet for a given group.}
    \label{tab:cultural_facet_accuracy}
\end{table}

We breakdown the performance of our highest performing model, Llama3-70B, for Prompt 1,\footnote{ Prompt 1 is described in appendix~\ref{sec:appendix_prompts}} by cultural facets as shown in the Table~\ref{tab:cultural_facet_accuracy}. We focus our analysis on Llama3-70B because it is one of the most recent and advanced models at the time of this writing, demonstrating a lower preference for distractors. For this, we record a 74.29\% preference for the US Correct answers in the US Specified Setting and a 55.42\% for the Ghana correct answers in the Ghana Specified Setting.
Our analysis shows that Llama3-70B is most familiar with the `Food and Drinks' cultural facet in the US contexts, as it answers 14 out of 16 questions correctly, and `Geography' in the Ghana context, with 7 out of 9 correct answers. For `Social Customs and Lifestyle,' the model answered 83 out of 118 questions correctly (70.34\%), compared to 62 out of 118 for Ghana (52.54\%), indicating better familiarity with US social customs. 

In `Architecture/Housing,' it answered 16 out of 19 correctly for the US (84.21\%) compared to 8 out of 19 for Ghana (42.11\%), showing a stronger grasp for US culture.
 `Food and Drinks' being the most familiar facet in the US contexts may be due to enough representation of this facet in the training data from different datasets, while the strong familiarity in Ghana `Geography,' might stem from the stable and consistently documented nature of geographical information, which is less culturally variable and easier for the model to learn. Also, the model struggled the most with `Social Customs and Lifestyle' in both contexts, likely due to the intricacies and nuances of social customs, which vary widely. These speculations highlight the importance of balanced training datasets to improve model performance across different cultural facets.

\section{Related Work}
\paragraph{Data Collection and Annotation}
Researchers often rely on collaborative efforts from diverse cultures at different stages in their work. \citet{huang-etal-2019-cosmos} and \citet{clark-etal-2020-tydi} allow diverse annotators to create commonsense questions, ensuring real-world relevance and promoting natural question generation.
\citet{ponti_xcopa_2020} strategically recruits translators to make cultural and linguistic adjustments, ensuring high-quality translations. Our work builds on this to allow broader participation, benefiting from diverse multi-cultural perspectives. While participants do not explicitly create the initial questions, they have the freedom to suggest refinements to questions and contexts, ensuring cultural accuracy and relevance. 
\citet{zellers-etal-2018-swag} uses adversarial methods to generate multiple-choice answers validated by crowd workers. \citet{yin-etal-2021-broaden} employs human annotators for questions and correct answers, with automated distractors generation. Our approach involves annotators creating, selecting, and validating answers.
\paragraph{Cultural Sensitivity and Bias Detection in Commonsense Reasoning}
\citet{palta_fork_2023}, \citet{zhou_cultural_2023}, \citet{huang-yang-2023-culturally}, \citet{bauer-etal-2023-social} and \citet{naous-etal-2024-beer} evaluate reasoning through downstream QA tasks, inference, entity recognition and sentiment analysis tasks, and discuss Western cultural biases. \citet{shi2024culturebank} fine-tune LLMs on a dataset of self-narratives from TikTok, and measure cultural awareness through automatic entailment scores and human evaluations.
\citet{li2024culturegen} prompts models with culturally relevant sentences and analyses the text for cultural symbols and biases, focusing on diversity and implicit Western biases. \citet{rao2024normad} assesses LLM responses to social acceptability questions across cultures, revealing difficulties with non-Western norms.
Our study evaluates cultural adaptability with our test set that contains both explicit and implicit cultural references. We assess how well models can interpret and respond to culturally nuanced questions in a MCQ task.

\section{Conclusion}
We create a new test set to investigate cultural biases in models, focusing on Ghana and the US. The focus on the Ghanaian population, a harder-to-study yet important group, underlines the importance of representing diverse cultures in computational research. Our study offers multi-stage annotator participation throughout the dataset creation, ensuring we capture nuanced cultural specificities. Our experiments show that models favor US culture in unspecified cultural contexts and specifying cultural cues improves accuracy for both groups, albeit, the improvement is more pronounced for the US. These findings highlight the need for ongoing efforts to make models more culturally inclusive, particularly for low-resource languages. Mitigation techniques should involve annotators from diverse backgrounds at all stages of dataset creation, and training models tailored to specific cultures.

\section*{Limitations}

\textbf{Limited dimensions of culture}
Culture as a social concept, spans a wide range of practices, beliefs, and expressions. Our research focuses on specific cultural facets for only two select cultural groups, which represents just a small aspect of culture. This focus was necessary due to the scope of this study and the resource constraints our study presents. Future research could expand on this by exploring more cultural dimensions for more low-resource cultures, that will offer a more comprehensive understanding of diverse human cultures in language models.

\textbf{Cultural stereotype }
There is a risk that some survey questions inadvertently perpetuate cultural stereotypes, particularly towards some ethnic groups, regions, or states. For instance, questions that reference a specific aspect of an ethnic group's food culture, or social custom, that is well-known among other people in the country could create an implicit marker that primes participants to select a correct cultural context answer but potentially reinforce stereotypes about the ethnic group. 

\textbf{Incomplete representation of ethnic/region demographics }
The survey might overemphasize the opinions of the majority ethnic groups. This could occur if minority viewpoints are less recognized, leading to the majority's perspective being perceived as the ``correct'' answer.  
As such, our best-voted answers might represent the viewpoints of the majority group while other good but least-voted answers although relatable to other groups are in the minority. 

\textbf{Limited dataset size and skewed demographics}
The size of our dataset is limited because we had to rely on volunteer annotators, as we could not use standard crowdsourcing tools for the Ghanaian population. This population is harder to study but important to understand due to its unique cultural context. Additionally, the participants we recruited skewed towards younger subjects, which may not fully represent the broader demographic distribution of the populations studied. This age skew could influence the cultural insights and relevance of the responses collected. Future research should aim to include a more diverse age range and employ strategies to increase participation from underrepresented groups.
\section*{Ethics Statement}
Our research investigates cultural biases in commonsense reasoning models by exploring specific cultural facets. We recognize that exploring cultural nuances presents inherent risks, particularly in inadvertently simplifying or misrepresenting cultures. Our work which focused on two select cultures, inherently limits the breadth of cultural diversity and understanding. We involved annotators from diverse backgrounds from these two groups to validate our findings, but we acknowledge that this does not fully encompass the rich diversity within each culture. We also emphasize that while our findings may highlight the presence of cultural biases in certain models, it is crucial to approach the interpretation of our results with an understanding of their contextual limitations and the complexities of cultural representation in AI.
Finally, we utilized AI assistants in our experiments (e.g., paraphrasing prompts, etc.).
\section*{Acknowledgements}
We thank the anonymous reviewers for their constructive feedback.
We also thank Kwesi Cobbina for their invaluable insights, as well as Nelson Padua-Perez, Julius Amegadzie, Belinda Laryea, Deborah Laryea, Amanda Acquaye and Gatha Adhikari for their support in recruiting annotators. We extend our heartfelt gratitude to the volunteers from Academic City University College, the University of Maryland, as well as all other volunteers from other institutions who contributed to this work. Their support was instrumental in the successful completion of our project. This work was supported by NSF CAREER Award No. 2339746. Any opinions, findings, and conclusions or recommendations expressed in this material are those of the author(s) and do not necessarily reflect the views of the National Science Foundation.

\bibliography{anthology,custom}

\appendix
\section{Dataset Statistics}
\label{sec:dataset_stats}
\begin{table}[!ht]
    \centering
    \begin{tabular}{lrrr}
        \toprule
        \textbf{Dataset} & \textbf{No. of samples}  \\
        \midrule
        CSQA & 77 \\
SIQA & 53 \\
PIQA & 70 \\
        \bottomrule
    \end{tabular}
    \caption{Information showing the number of samples
used from each dataset}
    \label{tab:og_div}
\end{table}

\begin{table}[!ht]
    \centering
    \begin{tabular}{lrrr}
        \toprule
        \textbf{Disambiguation Type} & \textbf{No. of samples}  \\
        \midrule
        Explicit &  200\\
Implicit     & 150 \\
        \bottomrule
        Unspecified     & 175 \\
        \bottomrule
    \end{tabular}
    \caption{Information showing the number of samples for each disambiguation type. Unspecified types do not have any cultural reference(either overt or subtle). }
    \label{tab:disamb}
\end{table}
\begin{table}[!ht]
    \centering
    \resizebox{0.78\linewidth}{!}{
    \begin{tabular}{lrrr}
        \toprule
        \textbf{Cultural Facet} & \textbf{No. of samples}  \\
        \midrule
        Social Customs and Lifestyle & 354 \\
Architecture/Housing & 57 \\
Food and Drinks & 48 \\
Gender, Marriage and Family & 39 \\
Geography & 27 \\

        \bottomrule
        Total & 525 \\
        \bottomrule
    \end{tabular}
    }
    \caption{Test set breakdown for different cultural facets.}
    \label{tab:cultural_facet}
\end{table}

\section{Prompts}
\label{sec:appendix_prompts}
Three prompts were used, increasing in specificity and detail from Prompt 1 to Prompt 3, referencing specific contexts and settings.

\subsection{Prompt 1}
\begin{figure}[H]
    \centering
    \noindent\fbox{%
        \parbox{\columnwidth}{%
        \scriptsize
        "You will be tasked with answering Multiple Choice Questions (MCQs). For each question, identify the answer that best fits the given context. Despite multiple options potentially being correct, select the one that is most appropriate for the specific scenario presented. Your response should be formatted in JSON, indicating the 'Best Answer' along with its corresponding option number and text."
        }%
    }
    \caption{General MCQ task with no specific context provided.}
    \label{fig:prompt1}
\end{figure}

\subsection{Prompt 2}
\begin{figure}[H]
    \centering
    \noindent\fbox{%
        \parbox{\columnwidth}{%
        \scriptsize
        "You will be tasked with answering Multiple Choice Questions (MCQs). For each question, identify the answer that best fits the given context for \textit{\textbf{a/an [specific setting]}} setting. Your response should be formatted in JSON, indicating the 'Best Answer' along with its corresponding option number and text."
        }%
    }
    \caption{MCQ task with a specific setting provided.}
    \label{fig:prompt2}
\end{figure}

\subsection{Prompt 3}
\begin{figure}[H]
    \centering
    \noindent\fbox{%
        \parbox{\columnwidth}{%
        \scriptsize
        "You will be tasked with answering Multiple Choice Questions (MCQs). For each question, answer as \textit{\textbf{a/an [specific culture]}} the answer that best fits the given context for \textbf{\textit{a/an [specific setting]}}. Your response should be formatted in JSON, indicating the 'Best Answer' along with its corresponding option number and text."
        }%
    }
    \caption{MCQ task with both a specific cultural perspective and setting provided.}
    \label{fig:prompt3}
\end{figure}

\section{Additional Results}
\label{sec:additional_results}
\begin{table*}[]
\centering
\resizebox{0.9\linewidth}{!}{
    \begin{tabular}{@{}l|llll|llll|llll@{}}
    \toprule
                     & \multicolumn{4}{c|}{\textbf{Answer-Only}}   & \multicolumn{4}{c|}{\textbf{\text{Question-and-Answers}}}   & \multicolumn{4}{c}{\textbf{($\Delta = \text{Question-and-Answers} - \text{Answer-Only}$)}} \\ 
                     \midrule
                     & \textbf{\%GH} & \textbf{\%US} & \textbf{\%GH Dist.} & \textbf{\%US Dist.} & \textbf{\%GH} & \textbf{\%US} & \textbf{\%GH Dist.} & \textbf{\%US Dist.} & \textbf{\%GH} & \textbf{\%US} & \textbf{\%GH Dist.} & \textbf{\%US Dist.} \\
                     \midrule
   bert-base        & 31.43         & 32.57         & 17.14           & 18.86         & 32.00          & 45.14          & 15.43             & 7.43           & \textcolor{blue}{+0.57}  & \textcolor{blue}{+12.57}  & \textcolor{red}{-1.71}  & \textcolor{red}{-11.43} \\ 
   \midrule
    roberta-base     & 25.14         & 35.43         & 18.29           & 21.14         & 30.29          & 51.43          & 10.28             & 8.00           & \textcolor{blue}{+5.15}  & \textcolor{blue}{+16.00}& \textcolor{red}{-8.01}  & \textcolor{red}{-13.14} \\  

   \midrule
    Llama2-7B       & 29.86& 24.91& 24.91& 20.34& 36.77          & 44.38          & 10.65& 8.20           & \textcolor{blue}{+6.91}& \textcolor{blue}{+19.47}& \textcolor{red}{-14.26}& \textcolor{red}{-12.14}\\ 
    Llama2-13B      & 32.62& 30.90& 21.19& 15.29& 30.05& 51.96& 10.05& 7.94& \textcolor{red}{-2.57}& \textcolor{blue}{+21.06}& \textcolor{red}{-11.14}& \textcolor{red}{-7.35}\\ 
    Llama2-70B& 40.47& 39.14& 10.39&  10.00& 29.95& 59.48& 6.34& 4.23& \textcolor{red}{-10.52}& \textcolor{blue}{+20.34}& \textcolor{red}{-4.05}& \textcolor{red}{-5.77}\\ 
    Llama3-70B& 41.77& 41.58& 9.95& 6.70& 23.43& 68.57& 5.33& 2.67& \textcolor{red}{-18.34}& \textcolor{blue}{+26.99}& \textcolor{red}{-4.62}& \textcolor{red}{-4.03}\\ 
   \midrule
    Gemma-7B        & 34.86         & 33.34         & 16.57& 15.23& 27.95          & 57.09          & 7.19              & 7.77& \textcolor{red}{-6.91}   & \textcolor{blue}{+23.75}  & \textcolor{red}{-9.38}& \textcolor{red}{-7.46}\\ 
   \midrule
    Mistral-7B      & 38.08         & 35.99         & 11.16           & 14.77& 30.66          & 60.19          & 4.00              & 5.14           & \textcolor{red}{-7.42}   & \textcolor{blue}{+24.20}  & \textcolor{red}{-7.16}  & \textcolor{red}{-9.63}\\
    \bottomrule              
    \end{tabular}
}
\caption{Models Preference Across Unspecified Cultural Settings. The performance of the models in the \text{Answer-Only} evaluates a model preference without any cultural context or questions provided. Columns "\%GH" and "\%US" show the percentage of correct Ghanaian and US answers, "\% GH Dist." the percentage of Ghanaian distractors, and "\% US Dist." the percentage of US distractors. The \text{Question-and-Answers} columns show similar metrics with unspecified questions provided. The ($\Delta = \text{Question-and-Answers} - \text{Answer-Only}$)
column shows the actual differences between the corresponding values in the Answer-Only and \text{Question-and-Answers} columns.}
\label{tab:unspecified_res}
\end{table*}
\begin{table}[!ht]
\centering
\resizebox{\linewidth}{!}{ 
\footnotesize{
\begin{tabular}{@{}l|llll|llll@{}}
\toprule
                 & \multicolumn{4}{c|}{\textbf{GH Specified}} & \multicolumn{4}{c}{\textbf{US Specified}} \\
\midrule
                 & \textbf{GH} & \textbf{\%US} & \textbf{\%GH Dist.} & \textbf{\%US Dist.} & \textbf{GH} & \textbf{\%US} & \textbf{\%GH Dist.} & \textbf{\%US Dist.} \\
\midrule
bert-base        & 34.0 & 37.0 & 18.0 & 11.0 & 28.0 & 44.0 & 15.0 & 13.0 \\
roberta-base     & 33.0 & 46.0 & 14.0 & 7.0 & 31.0 & 52.0 & 10.0 & 7.0 \\ \midrule
Llama2-7B        & 38.0 & 37.0 & 15.0 & 10.0 & 26.0 & 53.0 & 10.0 & 11.0 \\
Llama2-13B       & 42.25& 37.25& 9.25& 11.25& 29.0 & 54.0 & 7.5& 9.5\\
Llama2-70B       & 55.0 & 35.0 & 6.0 & 4.0& 23.0 & 67.0 & 4.0& 6.0 \\
Llama3-70B       & 60.0 & 33.0 & 4.1& 2.9& 18.0 & 77.0 & 3.0& 2.0\\ \midrule
Gemma-7B         & 50.0 & 34.25& 8.75& 7.0 & 24.25& 62.0 & 5.0 & 8.75\\ \midrule
Mistral-7B       & 48.0 & 42.0 & 5.0 & 5.0 & 25.0 & 65.0 & 5.0& 5.0 \\
\bottomrule

\end{tabular}
} }

\caption{Full Results on Model Preference Distribution in GH and US Specified Settings. Columns "\%GH" and "\%US" show the percentage of correct Ghanaian and US answers, "\%GH Dist." the percentage of Ghanaian distractors, and "\%US Dist." the percentage of US distractors. }
\label{tab:full_GH_US}
\end{table}

\begin{table}[!ht]
\centering
\resizebox{\linewidth}{!}{ 
\footnotesize{
\begin{tabular}{@{}l|ll|ll|ll@{}}
\toprule
                 & \multicolumn{2}{c|}{\textbf{UN Specified}} & \multicolumn{2}{c|}{\textbf{GH Specified}} & \multicolumn{2}{c}{\textbf{US Specified}} \\
\midrule
                 & \textbf{\%GH} & \textbf{\%US} & \textbf{\%GH} & \textbf{\%US} & \textbf{\%GH} & \textbf{\%US} \\
\midrule
bert-base        & 41.0 & 59.0 & 47.0 & 53.0 & 41.0 & 59.0 \\
roberta-base     & 37.0 & 63.0 & 42.0 & 58.0 & 39.0 & 61.0 \\ \midrule
Llama2-7B        & 51.0 & 49.0 & 49.0 & 51.0 & 37.0 & 63.0 \\
Llama2-13B       & 35.0 & 65.0 & 58.0 & 42.0 & 31.0 & 69.0 \\
Llama2-70B       & 37.0 & 63.0 & 66.0 & 34.0 & 25.0 & 75.0 \\
Llama3-70B       & 26.0 & 74.0 & 67.0 & 33.0 & 18.0 & 82.0 \\ \midrule
Gemma-7B         & 40.0 & 60.0 & 59.0 & 41.0 & 30.0 & 70.0 \\ \midrule
Mistral-7B       & 34.0 & 66.0 & 56.0 & 44.0 & 25.0 & 75.0 \\
\bottomrule

\end{tabular}
} }

\caption{Full Results on Model Preference Distribution in UN, GH, and US Specified Settings for the Correct-only configuration. Columns "\%GH" and "\%US" show the percentage of correct Ghanaian and US answers.}
\label{tab:full_conditioned}
\end{table}




\begin{figure}[H]
     \centering
     \includegraphics[width=1\linewidth]{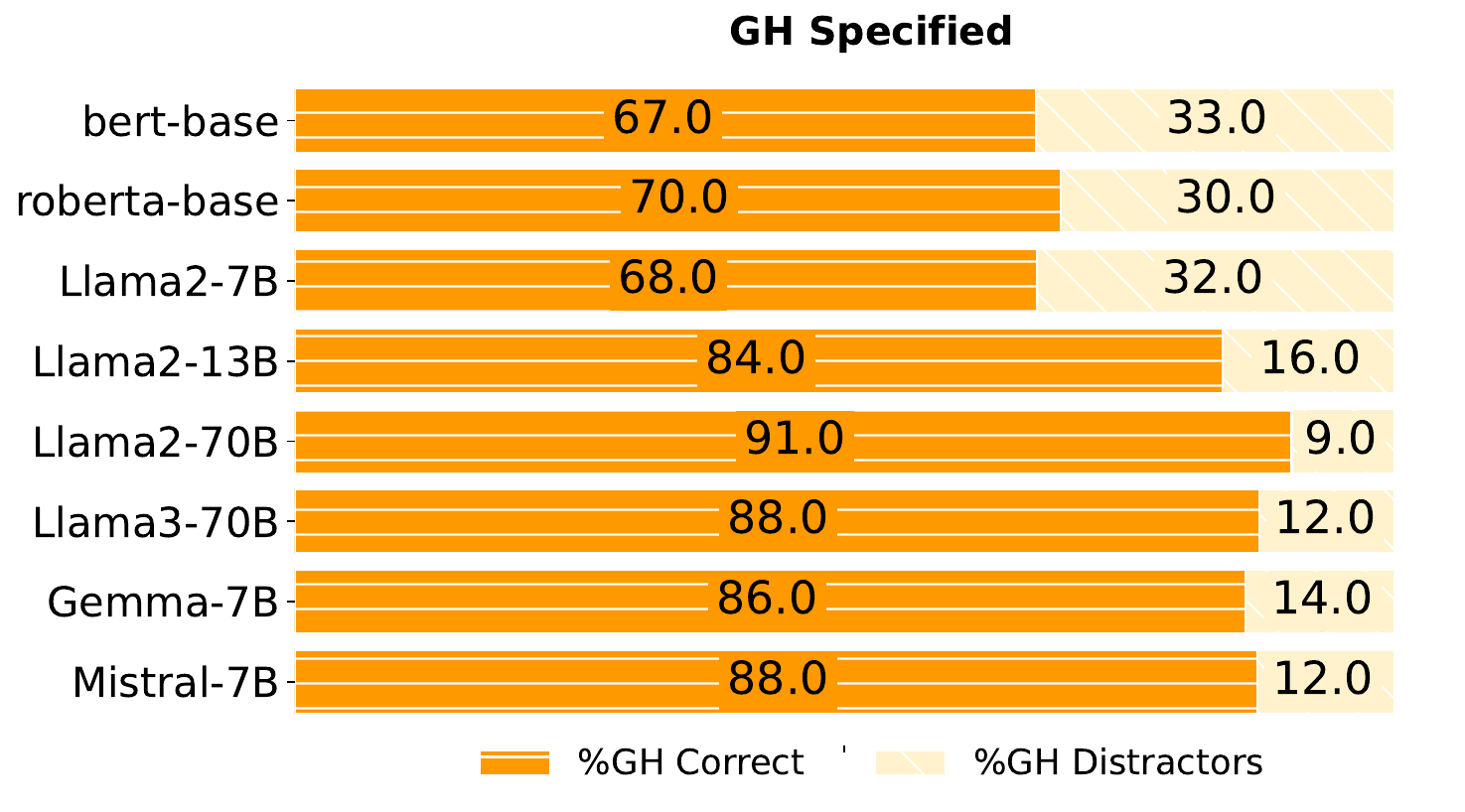}
     \caption{Accuracy of models in Ghanaian settings when conditioned only on Ghanaian correct answers and Ghanaian distractors.}
     \label{fig:gh_gh_only}
 \end{figure}
 \begin{figure}[H]
     \centering
     \includegraphics[width=1\linewidth]{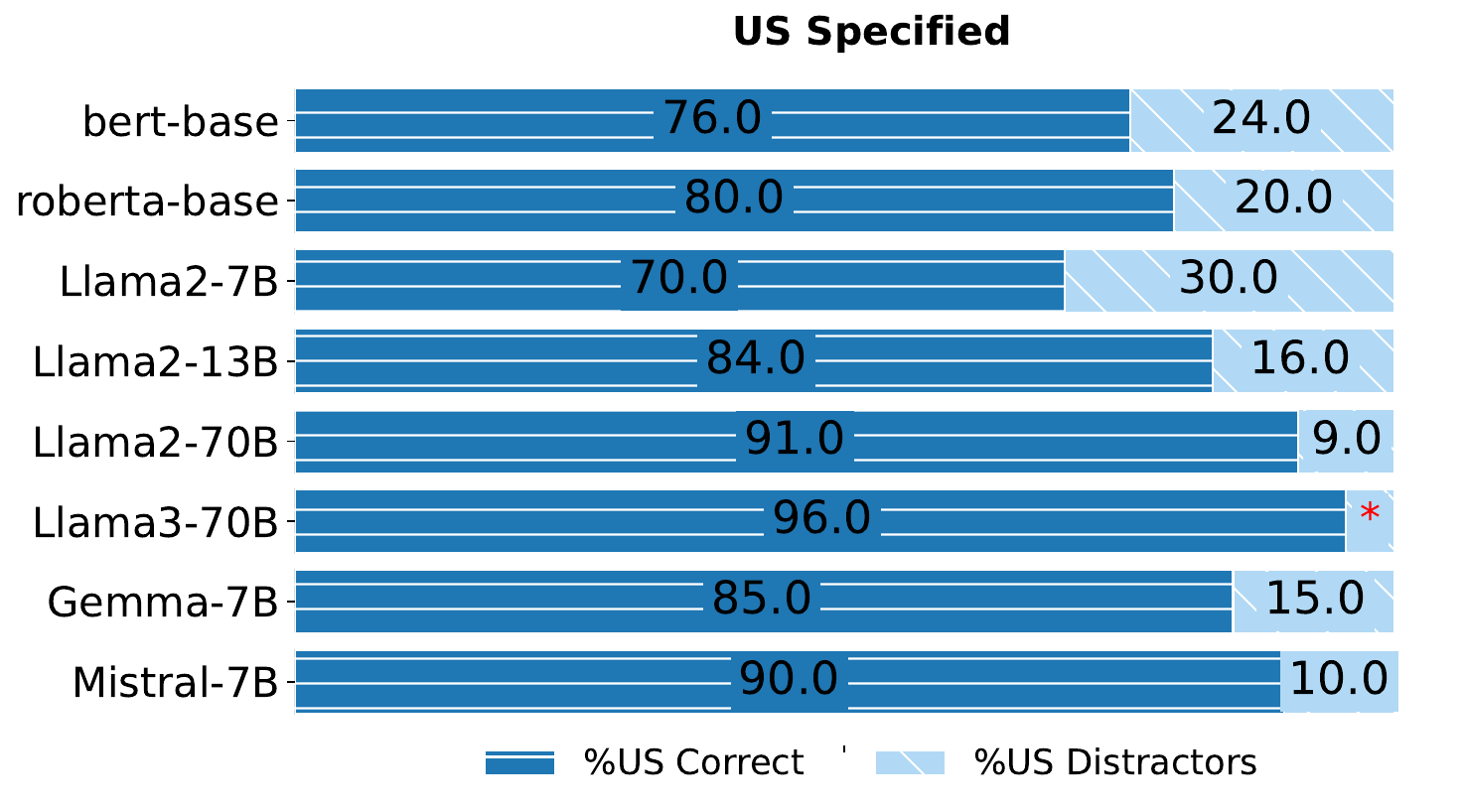}
     \caption{Accuracy of models in US settings when conditioned only on US correct answers and US distractors.}
     \label{fig:us_us_only}
 \end{figure}

\section{Qualitative Results}
\label{sec:qualitiative}
This section shows the different model responses for one GH-specified test question when all four answer choices are provided. 

\textbf{Model Responses:}
\vspace{0.5cm}

\noindent\fbox{%
    \begin{minipage}{\columnwidth}
    \textbf{bert-base:} 
    
    Decorate the Christmas tree with lots of presents underneath it, and decorate the house and the front porch with ornaments and lights
    \end{minipage}
}

\vspace{0.5cm}

\noindent\fbox{%
    \begin{minipage}{\columnwidth}
    \textbf{roberta-base:} 
    
    Decorate the Christmas tree with lots of presents underneath it, and decorate the house and the front porch with ornaments and lights
    \end{minipage}
}

\vspace{0.5cm}

\noindent\fbox{%
    \begin{minipage}{\columnwidth}
    \textbf{Llama2-7B:}
    \begin{quote}\scriptsize
    { "Best Answer": "(3) Ensure there is enough food, drinks, and fun games for the whole family.", \\
    "Explanation": "The context of the question is that Kojo wants to make Bronya feel more magical for his family, and the options provided are all related to Christmas celebrations. Option (3) is the best answer because it suggests ensuring that there is enough food, drinks, and fun games for the whole family, which is a great way to create a magical and enjoyable experience for everyone." }
    \end{quote}
    \end{minipage}
}

\vspace{0.5cm}

\noindent\fbox{%
    \begin{minipage}{\columnwidth}
    \textbf{Llama2-13B:}
    \begin{quote}\scriptsize
    { "Best Answer": "(1) Decorate the Christmas tree with lots of presents underneath it, and decorate the house and the front porch with ornaments and lights", \\
    "Explanation": "To make Bronya feel more magical for his family, Kojo should focus on creating a festive and joyful atmosphere around the Christmas tree. Decorating the tree with lots of presents underneath it, and the house and front porch with ornaments and lights, will help create a sense of wonder and excitement for the children. This option aligns best with the context of the question, which is focused on creating a magical atmosphere for the family during the holiday season." }
    \end{quote}
    \end{minipage}
}

\vspace{0.5cm}

\noindent\fbox{%
    \begin{minipage}{\columnwidth}
    \textbf{Llama3-70B:}
    \begin{quote}\scriptsize
    { "Best Answer": "(1) Decorate the Christmas tree with lots of presents underneath it, and decorate the house and the front porch with ornaments and lights", \\
    "Explanation": "This answer is the most suitable because it creates a festive and magical atmosphere for Kojo's family, especially his children, a key aspect of the Christmas season." }
    \end{quote}
    \end{minipage}
}

\vspace{0.5cm}

\noindent\fbox{%
    \begin{minipage}{\columnwidth}
    \textbf{Gemma-7B:}
    \begin{quote}\scriptsize
    { "Best Answer": "(3) Ensure there is enough food, drinks and fun games", \\
    "Explanation": "The text describes a person who is married with two little kids and wants to make Bronya feel more magical for their family." }
    \end{quote}
    \end{minipage}
}

\vspace{0.5cm}

\noindent\fbox{%
    \begin{minipage}{\columnwidth}
    \textbf{Mistral-7B:}
    \begin{quote}\scriptsize
    { "Best Answer": "1. Decorate the Christmas tree with lots of presents underneath it, and decorate the house and the front porch with ornaments and lights.", \\
    "Explanation": "The context mentions that Kojo has a family with two little kids. The question asks how Kojo can make Bronya feel more magical for his family. Option 1 is the best answer as it aligns with the context and the holiday season. Decorating the Christmas tree." }
    \end{quote}
    \end{minipage}
}
\section{ Additional Experimental Setup Details }
\label{sec:appendix_experiments}

\paragraph{Encoder Models}
 We finetuned the encoder models using a consolidated version of the original datasets,(CSQA, SIQA, and PIQA) for 3 epochs and a learning rate of 1e-5 for the base models. The consolidated dataset for the models was created by standardizing the number of answer choices across the different datasets to four answer choices per question. We evaluated these finetuned encoder models on the validation sets of these original datasets to ensure it is consistent with existing baselines. Performance-wise, on the consolidated dataset, BERT-base achieved scores of 60.77\% on CSQA, 60.29\% on SIQA, and 65.83\% on PIQA. Roberta-base performed better with scores of 69.37\% on CSQA, 70.37\% on SIQA, and 69.37\% on PIQA. 
 \paragraph{LLMs}
 For the Llama2 models, Gemma and Mistral, we run the experiments with a random seed of 42. We used a temperature of 0.1 for all LLMs. 
 \\
 \textbf{Summary of Experiments}
 \begin{table}[!h]
    \centering
    \resizebox{\linewidth}{!}{
    \begin{tabular}{lllp{8cm}}
        \toprule
        \textbf{Exp. Setting} & \textbf{Cultural Setting} & \textbf{Exp. Configuration} & \textbf{Answers Provided} \\
        \midrule
        Unspecified & No culture & All-answers & GH Correct, GH Dist., US Correct, US Dist. \\
        & No culture & Correct-only & GH Correct, US Correct \\ \midrule
        Specified & GH & All-answers & GH Correct, GH Dist., US Correct, US Dist. \\
        & GH & Correct-only & GH Correct, US Correct \\
       & GH & GH-only & GH Correct, GH Dist. \\ \midrule
        Specified & US & All-answers & GH Correct, GH Dist., US Correct, US Dist. \\
         & US & Correct-only & GH Correct, US Correct \\
         & US & US-only & US Correct, US Dist. \\ \midrule
        Answer-Only & No culture & All-answers & GH Correct, GH Dist., US Correct, US Dist. \\
        \bottomrule
    \end{tabular}
    }
    \caption{This table summarizes all experiments we conduct. GH = Ghana, Dist. = Distractors.}
    \label{tab:experiment_configurations}
\end{table}

\paragraph{Terms of use for each model}
We carefully follow the guidelines per the terms of usage described by the model authors or company

\begin{itemize}
    \item Llama2: \url{https://ai.meta.com/llama/license/}
    \item Llama3: \url{https://llama.meta.com/llama3/license/}
    \item Mistral-Instruct-v0.1: \url{https://mistral.ai/terms-of-service/}
    \item Gemma: \url{https://github.com/google-deepmind/gemma/blob/main/LICENSE}
\end{itemize}
\paragraph{Licenses}
The CSQA, SIQA and PIQA are used under the  MIT\footnote{\textcolor[HTML]{000099}{https://opensource.org/license/MIT}} and CC-BY\footnote{\textcolor[HTML]{000099}{https://creativecommons.org/licenses/by/4.0/}} licenses. 
\section{Disambiguation for the three datasets}
\label{sec:appendix_disambiguation_types}

    \begin{figure}[H]
        \centering
        \includegraphics[width=0.9\linewidth]{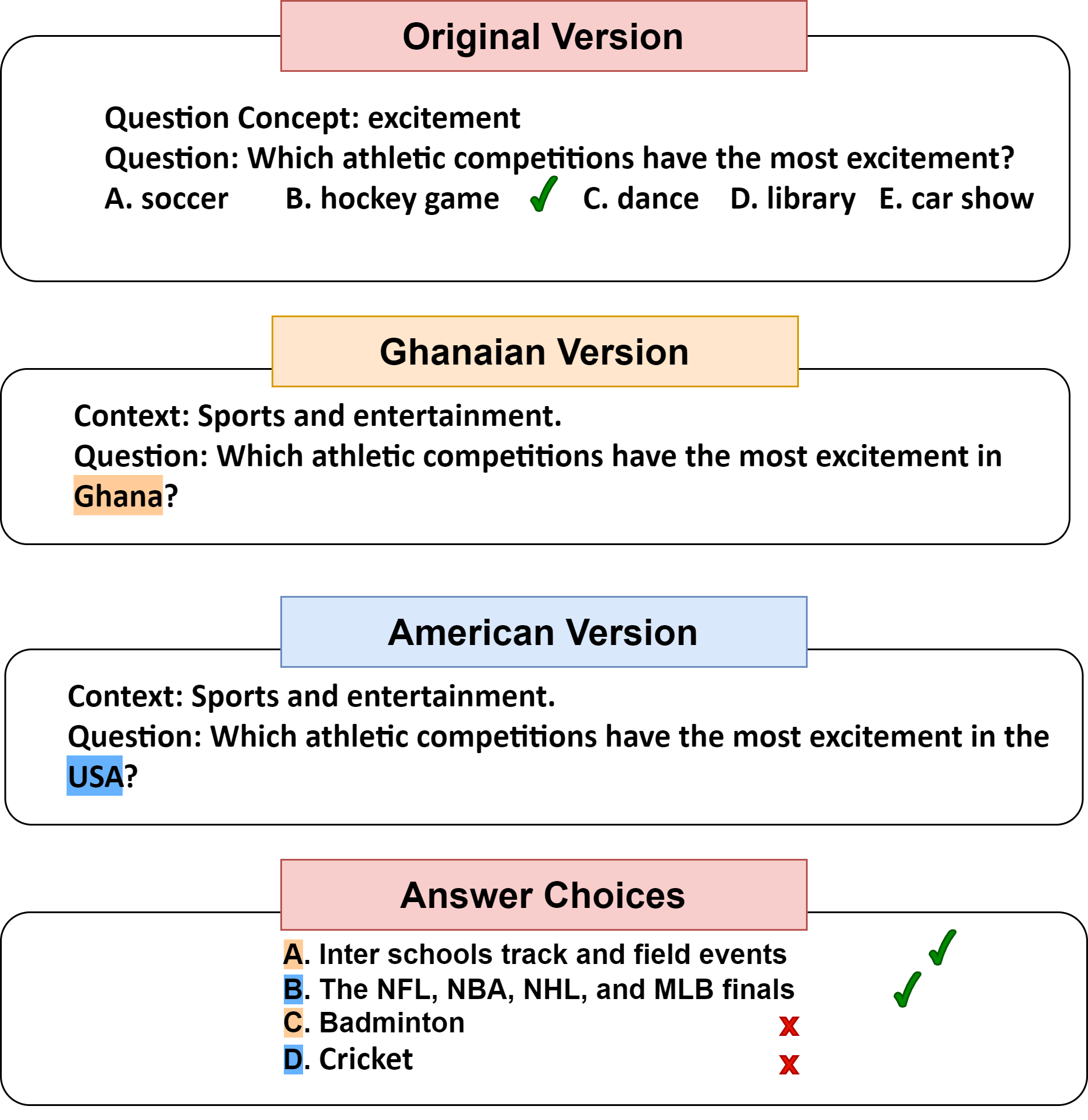}
        \caption{An example of the question-answer format in the original CSQA \cite{talmor-etal-2019-commonsenseqa} dataset and the disambiguated version from our dataset, with Ghanaian and American versions as well as the correct and distractor answers for both versions.}
        \label{fig:cqa}
    \end{figure}

\begin{figure}[!ht]
    \centering
    \includegraphics[width=1\linewidth]{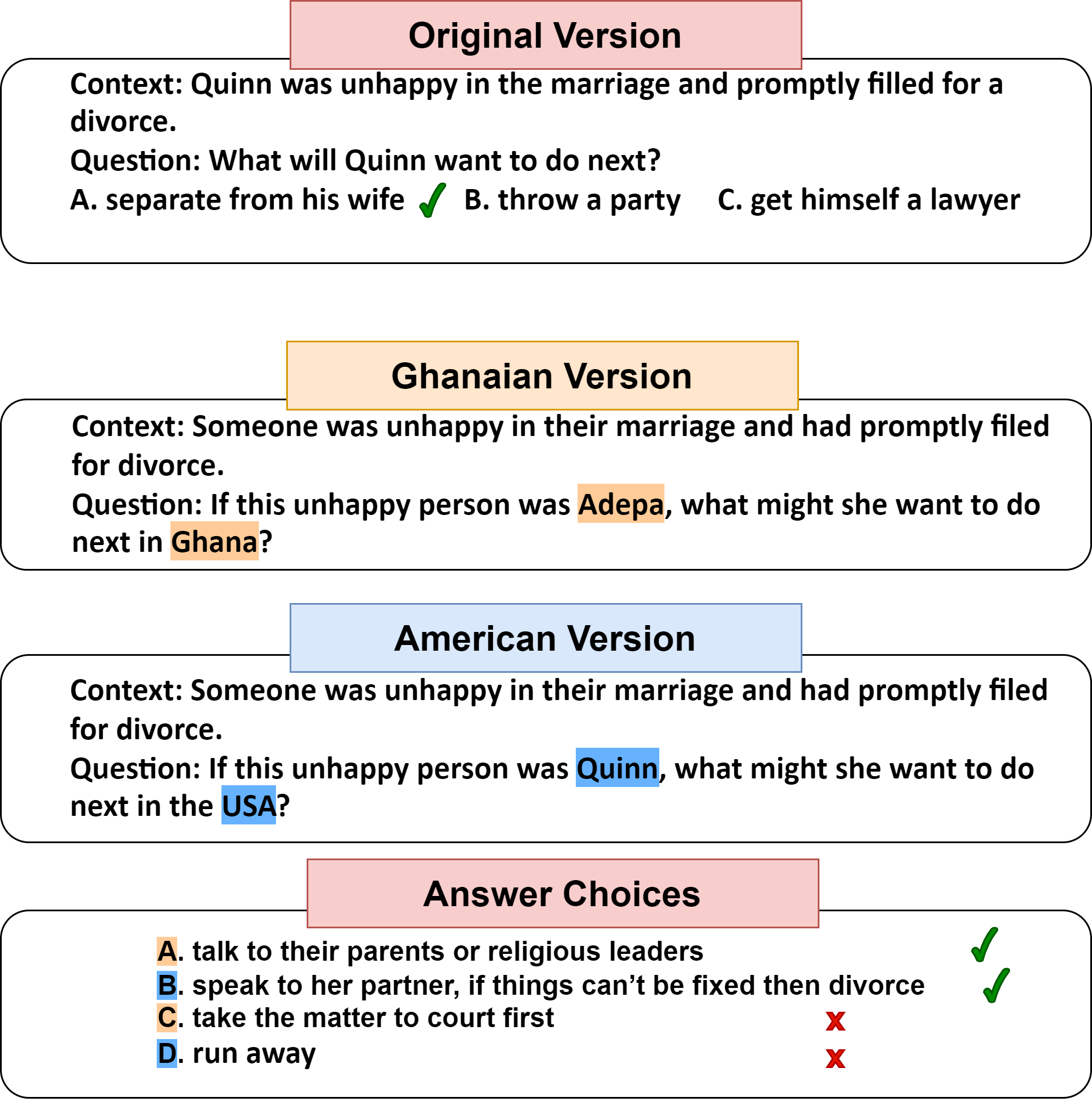}
    \caption{An example of the question-answer format in the original SIQA \cite{sap_social_2019} dataset and the disambiguated version from our dataset, with Ghanaian and American versions as well as the correct and distractor answers for both versions.}
    \label{fig:siqa}
\end{figure}

\begin{figure}[!ht]
    \centering
    \includegraphics[width=1\linewidth]{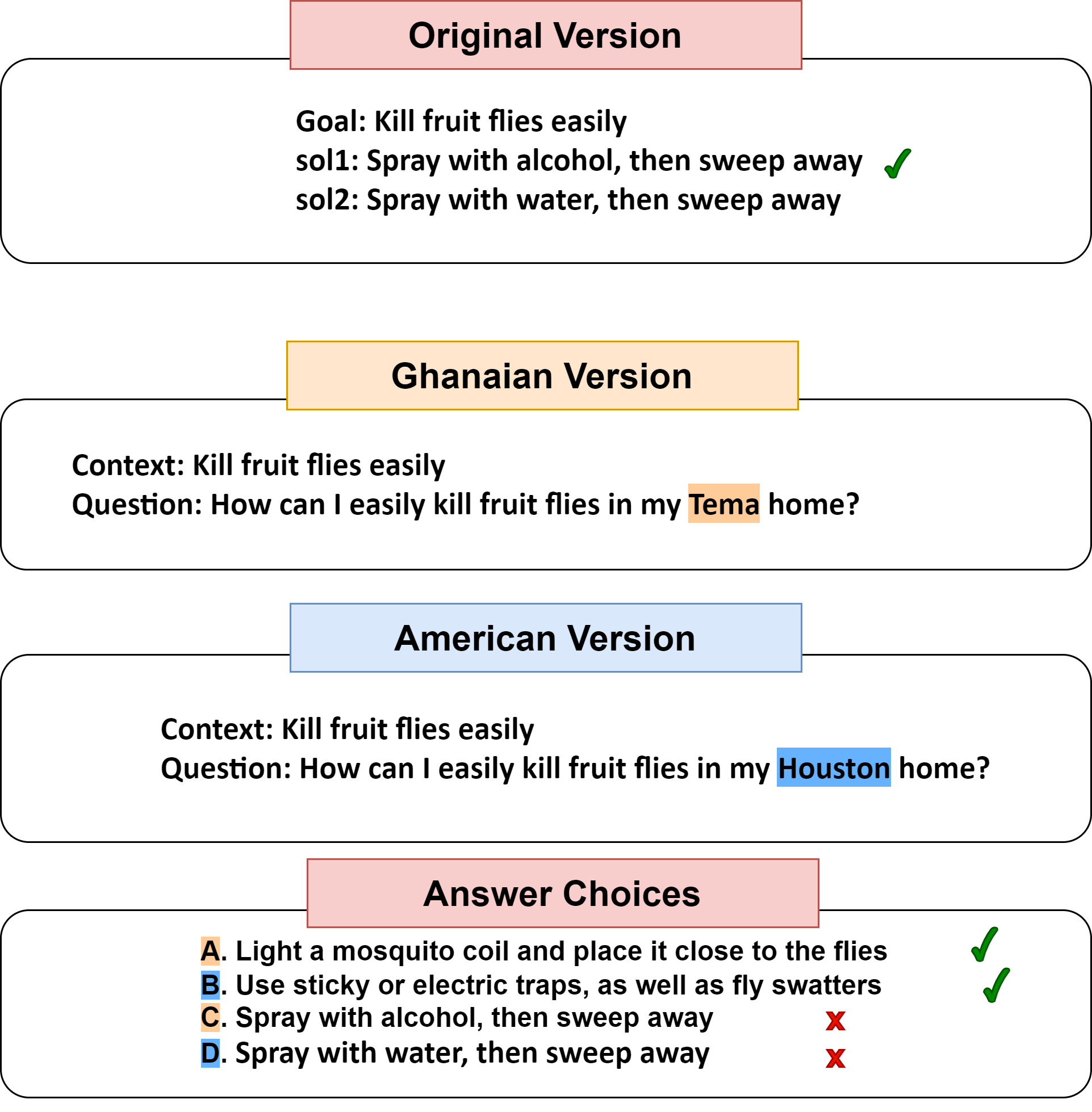}
    \caption{An example of the question-answer format in the original PIQA~\cite{Bisk_Zellers_Le_bras_Gao_Choi_2020} dataset and the disambiguated version from our dataset, with Ghanaian and American versions as well as the correct and distractor answers for both versions.} 
    \label{fig:piqa}
\end{figure}

\section{Surveys}
\label{sec:appendix_surveys}
\begin{figure}[H]
    \centering
    \includegraphics[width=0.9\linewidth]{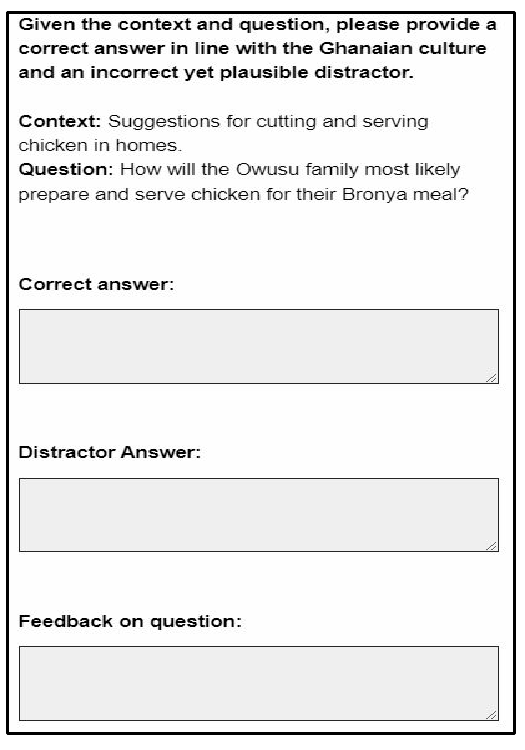}
    \caption{Survey Sample from the Answer Choice Generation Stage for Ghana}
    \label{fig:gh_1}
\end{figure}
\begin{figure}[H]
    \centering
    \includegraphics[width=0.9\linewidth]{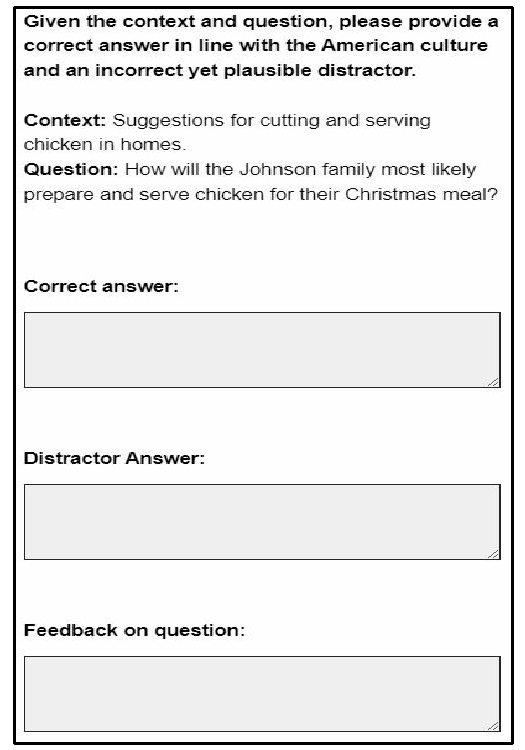}
    \caption{Survey Sample from the Answer Choice Generation Stage for the USA}
    \label{fig:us_1}
\end{figure}
\begin{figure}[H]
    \centering
    \includegraphics[width=0.9\linewidth]{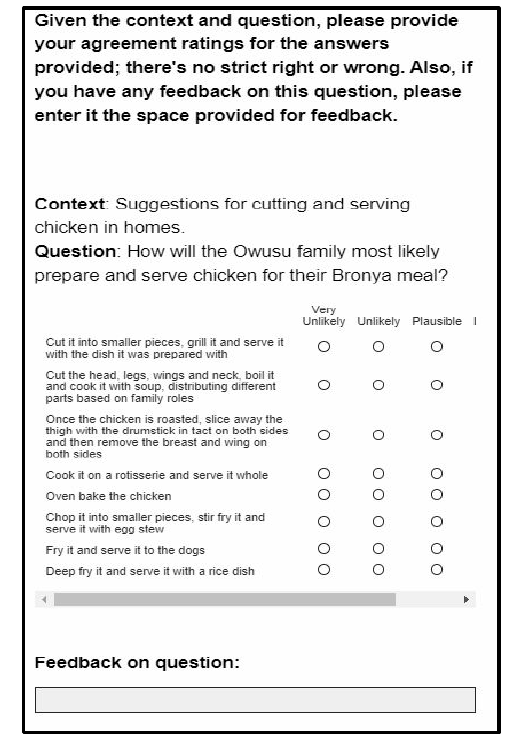}
    \caption{Survey Sample from the Likert Scale Answer Annotation Stage for Ghana}
    \label{fig:gh_2}
\end{figure}
\begin{figure}[H]
    \centering
    \includegraphics[width=0.85\linewidth]{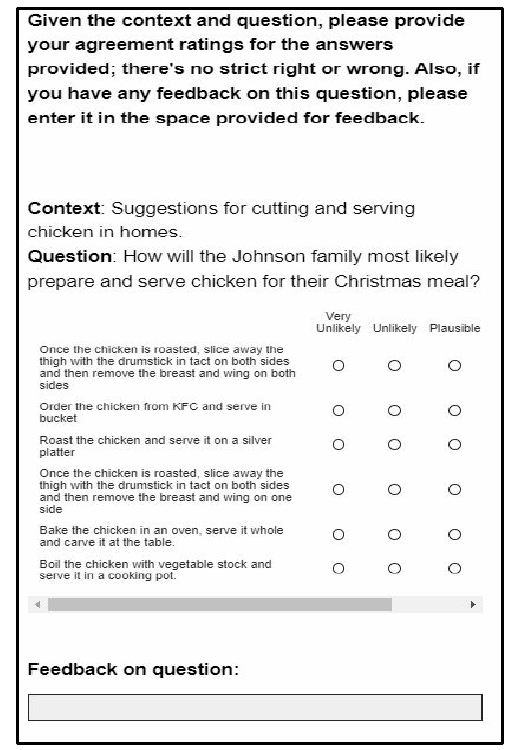}
    \caption{Survey Sample from the Likert Scale Answer Annotation Stage for the USA}
    \label{fig:us_}
\end{figure}
\begin{figure}[H]
    \centering
    \includegraphics[width=0.85\linewidth]{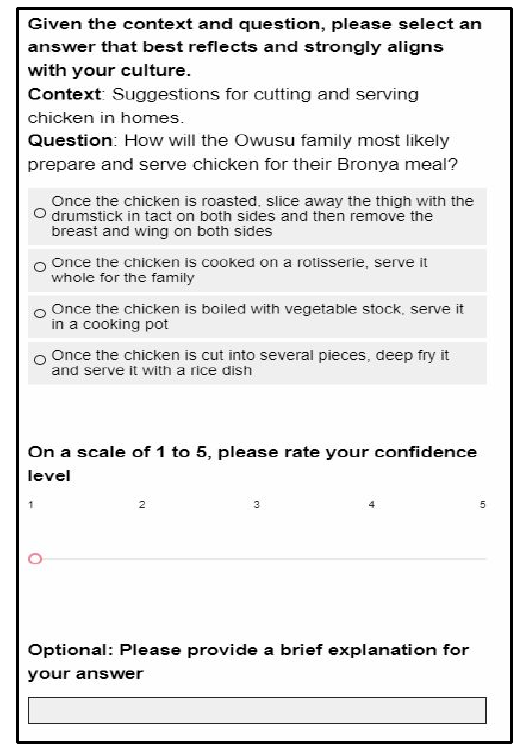}
    \caption{Survey Sample from the Multiple Answer Choice Annotation for Ghana. The slider scales for the confidence ratings were locked to integers, ensuring that respondents provided answers as whole numbers only.}
    \label{fig:gh_3}
\end{figure}
\begin{figure}[H]
    \centering
    \includegraphics[width=0.9\linewidth]{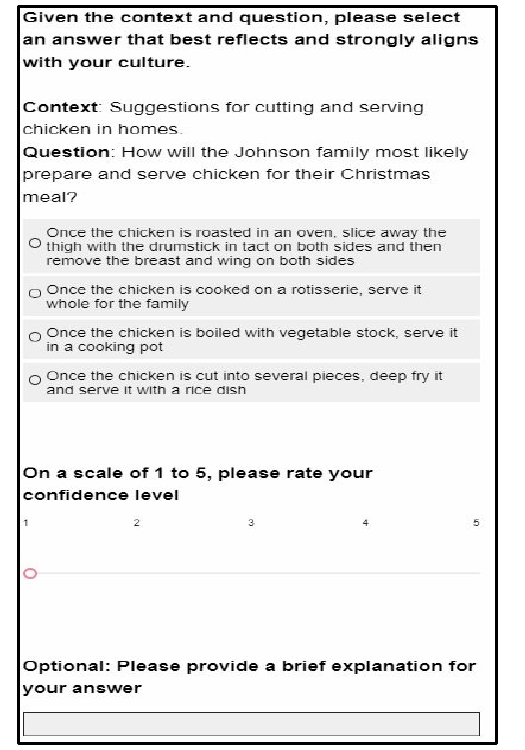}
    \caption{Survey Sample from the Multiple Answer Choice Annotation for the USA. The slider scales for the confidence ratings were locked to integers, ensuring that respondents provided answers as whole numbers only.}
    \label{fig:us_3}
\end{figure}

\section{Annotators}
\label{sec:appendix_annotators}
The annotators for each stage played a crucial role in validating the accuracy and relevance of these disambiguated questions and answers. 
\subsection{Annotator Elligibilty}
\label{sec:appendix_recruitment}
We recruited participants who were adults aged 18 years or older, fluent in English and self-identified as either American or Ghanaian. The eligibility criteria further inquired about the nationality (Ghanaian/American), current country of residence (Ghana/USA), and place of most time spent before turning 18 (Ghana/USA). These criteria ensured that participants were well situated and could provide culturally relevant responses, with annotators from the United States focusing on a U.S. cultural context and those from Ghana on a Ghanaian cultural context. 

\subsection{Recruitment Process}
Interested participants were invited to email their interest to a designated study-specific email address after which we present the full consent detailing the risks involved as well as other information relevant to the study. We responded to this interest email with a Qualtrics link for a particular survey set. We also obtained IRB approval for collecting, storing, and deleting the email addresses of participants. These email addresses were collected from direct email communications initiated by the participants themselves, who voluntarily reached out to express their interest in joining our study. The emails were used to track which survey set was assigned to each volunteer, ensuring that no participant received the same set in the same or different stages of the study.  We ensure that all participants give their consent to this by presenting it as the first question in our survey. We anonymize all responses before processing so the emails are not tracked in our dataset creation process.

This rigorous recruitment process ensured that our participants were adequately informed and consented to participate, aligning with ethical research standards.

\subsection{Annotator Demographics}

\label{sec:appendix_demographics}

Across all plots for both Ghanaian and American participants, we observe more participations in the Multiple Annotation stages. We present detailed demographic plots of our participants.These figures show some differences between Ghanaian and American annotators in terms of gender, age, education, ethnicity, and regional representation. Some observations include a higher number of male participants in the U.S. and a higher gender balance for Ghanaian annotators, with significant participation from younger individuals and varying educational backgrounds. 

\begin{figure*}[h]
    \centering
    \includegraphics[width=1\linewidth]{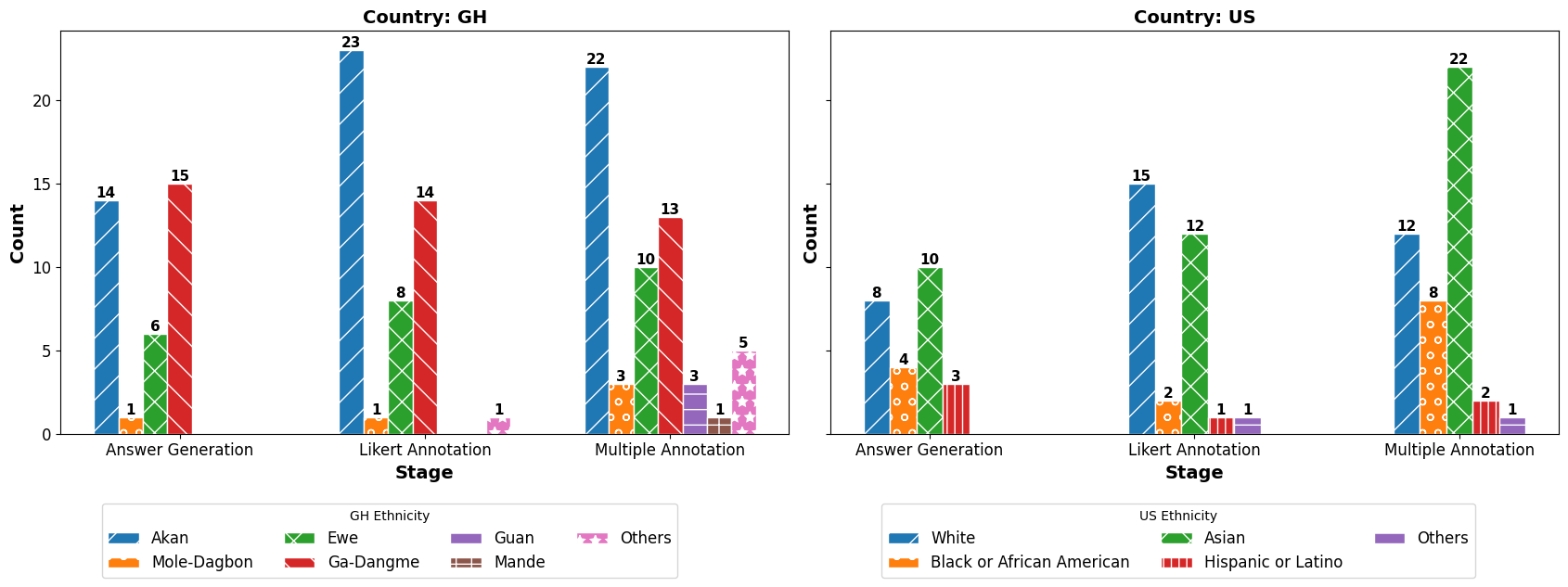}
    \caption{Ethnicity Distribution across Stages for Ghanaian and American Participants}
    \label{fig:ethnicitiy}
\end{figure*}

The ethnic group distribution for Ghanaians show more representation from the Akan and Ga-Dangme groups across all stages, while the distribution for Americans show more Asian and White participants across all stages. For detailed ethnicity breakdown, please refer to Figure~\ref{fig:ethnicitiy}.

\begin{figure*}[h]
    \centering
    \includegraphics[width=1\linewidth]{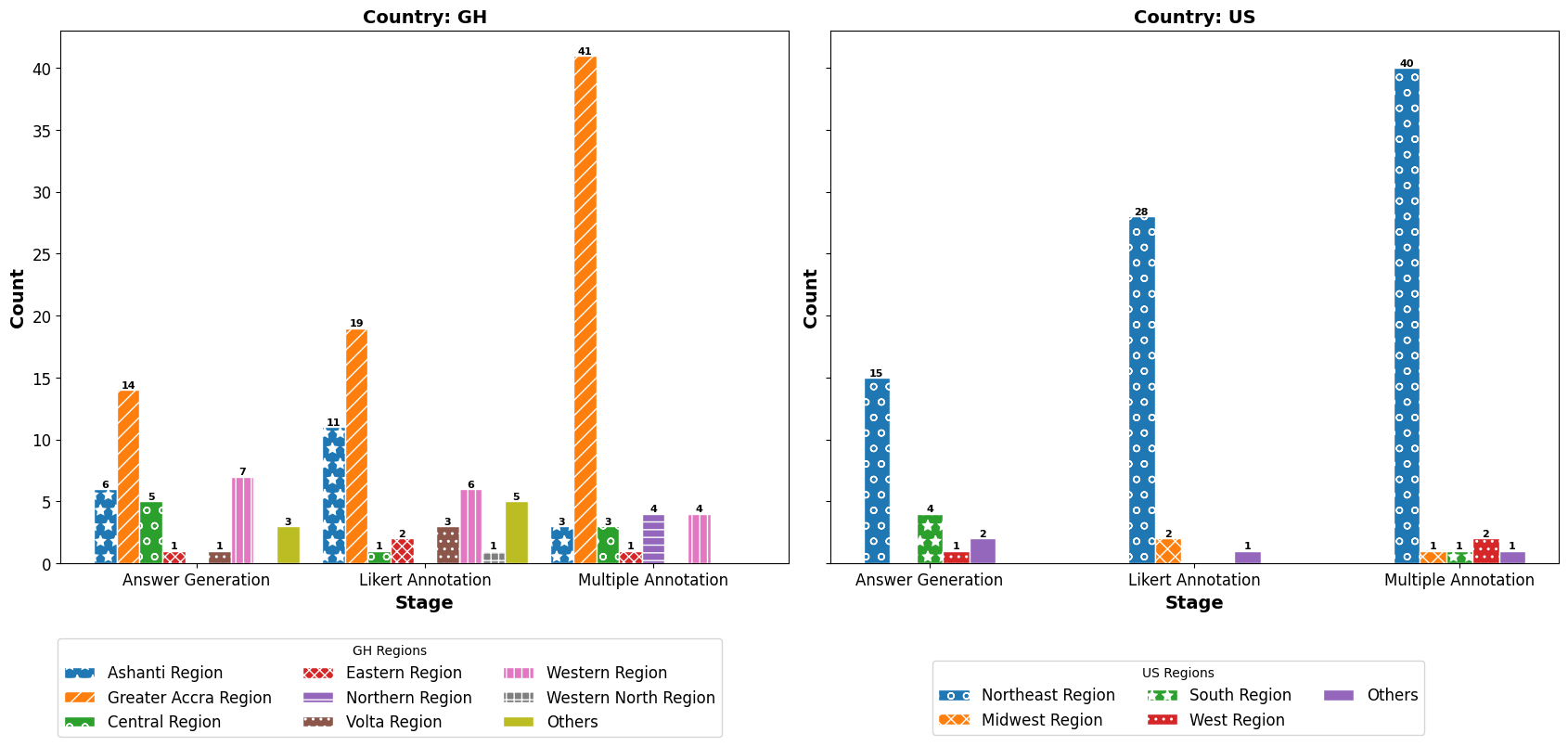}
    \caption{Region Distribution across Stages for  Ghanaian and American Participants}
    \label{fig:region}
\end{figure*}

Additionally, we see the regional distrbiton shows more representation from the Greater Accra Region across all stages, especially in the Multiple Answer Choice Annotation stage. For the American participants we also see more representation from the Northeast Region across all stages. Please refer to Figure~\ref{fig:region}.
\begin{figure*}[h]
    \centering
    \includegraphics[width=1\linewidth]{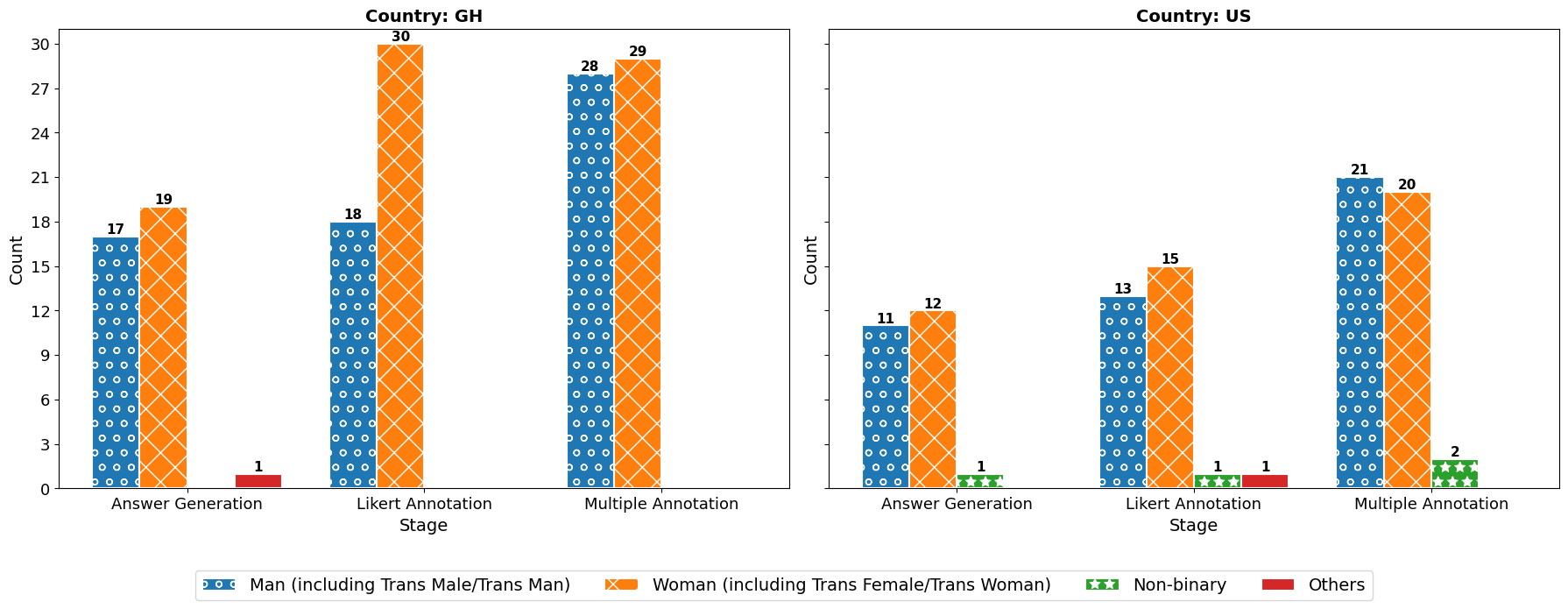}
    \caption{Gender Distribution across Stages for  Ghanaian and American Participants}
    \label{fig:gender}
\end{figure*}

\begin{figure*}[h]
    \centering
    \includegraphics[width=1\linewidth]{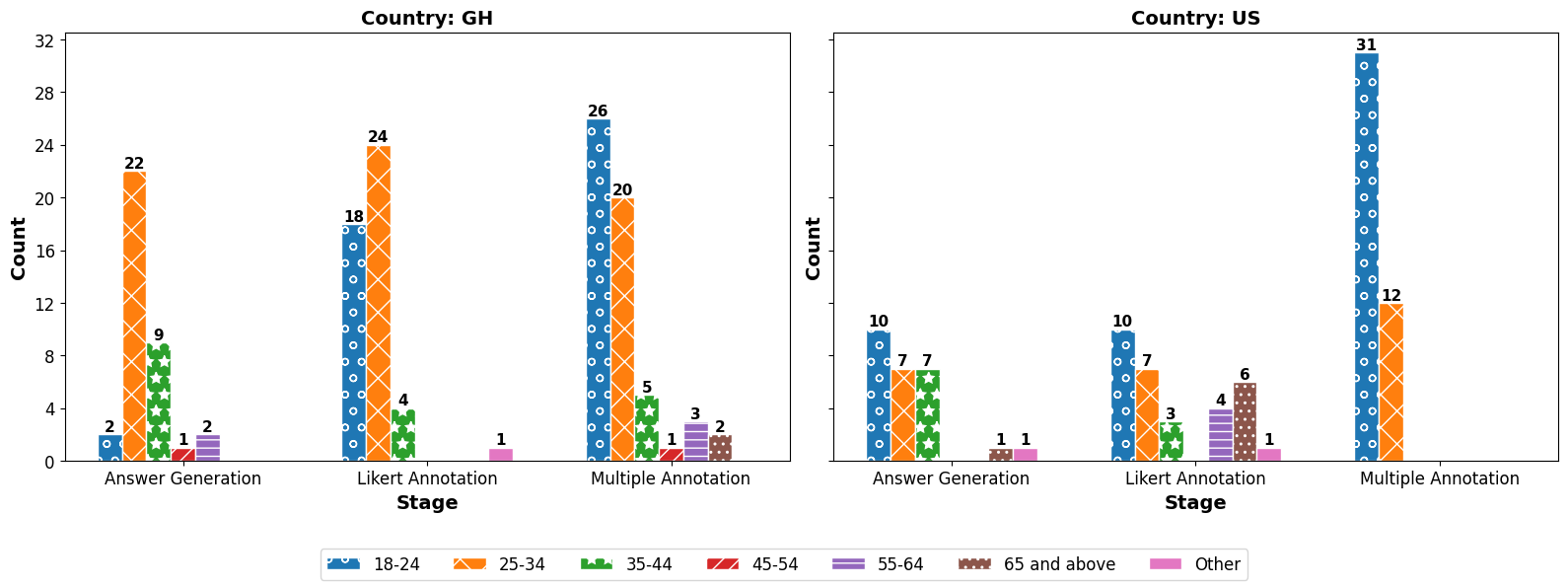}
    \caption{Age Distribution across Stages for Ghanaian and American Participants}
    \label{fig:age}
\end{figure*}

\begin{figure*}[h]
    \centering
    \includegraphics[width=1\linewidth]{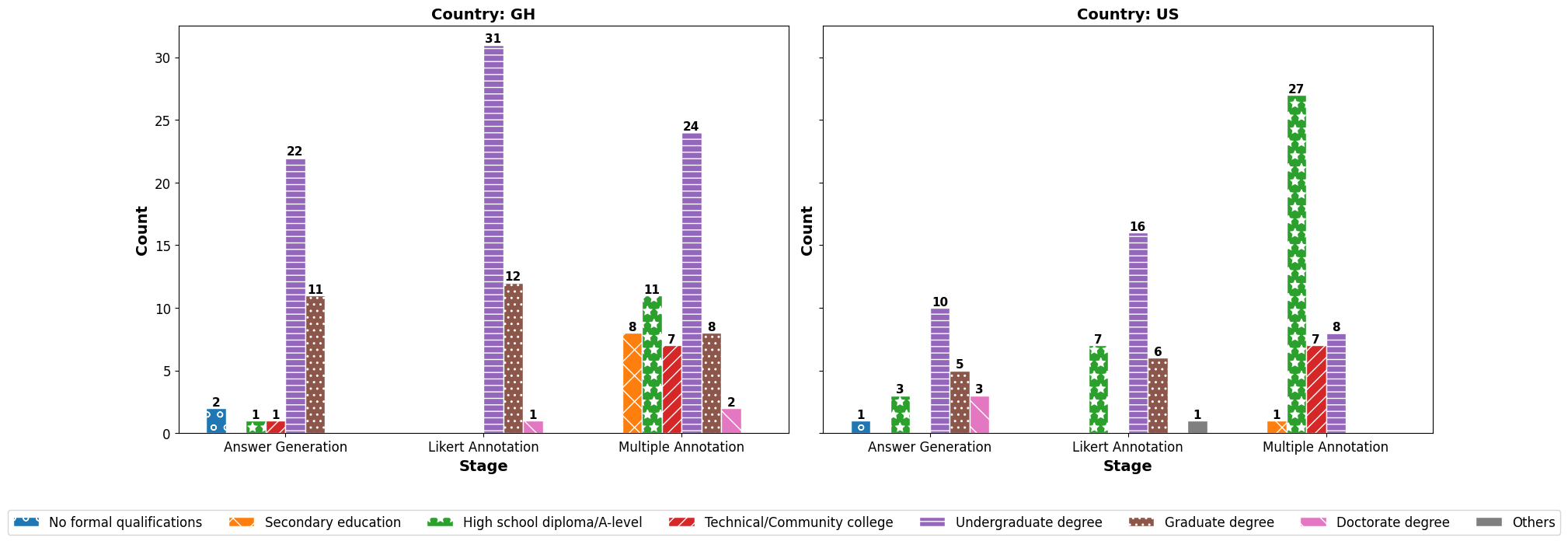}
    \caption{Education Distribution across Stages for  Ghanaian and American Participants}
    \label{fig:education}
\end{figure*}

\end{document}